\PassOptionsToPackage{sort}{natbib}
\documentclass[10pt,a4paper,authoryear]{nus-techreport}
\pdfoutput=1
\usepackage{tikz,needspace,wrapfig,url,placeins,nicematrix}
\usepackage[normalem]{ulem}
\usetikzlibrary{arrows.meta,positioning}
\usepackage{tabularx,pifont}
\newcolumntype{L}[1]{>{\hsize=#1\hsize\linewidth=\hsize\raggedright\arraybackslash}X}
\newcolumntype{C}[1]{>{\hsize=#1\hsize\linewidth=\hsize\centering\arraybackslash}X}
\newcolumntype{R}[1]{>{\hsize=#1\hsize\linewidth=\hsize\raggedleft\arraybackslash}X}
\definecolor{VitalInk}{HTML}{243247}
\definecolor{VitalBlue}{HTML}{2878B5}
\definecolor{VitalOrange}{HTML}{D47721}
\definecolor{VitalPurple}{HTML}{7950B5}
\definecolor{VitalLight}{HTML}{F4F6FA}
\definecolor{VitalRev}{HTML}{000000}
\DeclareRobustCommand{\method}{\textup{\textcolor{VitalBlue}{V}\textcolor{VitalBlue!75!VitalPurple}{I}\textcolor{VitalBlue!50!VitalPurple}{T}\textcolor{VitalBlue!25!VitalPurple}{A}\textcolor{VitalPurple}{L}}}
\hypersetup{colorlinks=true,citecolor=VitalBlue!85!VitalInk,
  linkcolor=VitalPurple!85!VitalInk,urlcolor=VitalBlue!85!VitalInk}
\newcommand{\vitalrow}{\rowcolor{VitalPurple!8}}
\newcommand{\vitalbest}[1]{\ifmmode\mathbf{#1}\else\textbf{#1}\fi}
\newcommand{\vitalsecond}[1]{\underline{#1}}
\newcommand{\independent}[1]{\par\noindent#1}
\newcommand{\rev}[1]{#1}
\newenvironment{revblock}{}{}
\DeclareRobustCommand{\vitaltablecitep}[1]{{\fontsize{6}{7}\selectfont\citep{#1}}}
\newcommand{\vitaltablefont}{\fontsize{7.2}{8.4}\selectfont}
\newcommand{\vitalup}{\ensuremath{\uparrow}}
\newcommand{\vitaldown}{\ensuremath{\downarrow}}
\DeclareMathSizes{7.2}{7.2}{5}{5}
\DeclareMathSizes{8.2}{8.2}{6}{5}
\makeatletter
\newcommand{\vitalfigurecaption}{\def\@captype{figure}\caption}
\newcommand{\vitaltablecaption}{\def\@captype{table}\caption}
\newcommand{\vitalappendixfloats}{%
  \raggedbottom
  \let\vitalmainmakecaption\@makecaption
  \long\def\@makecaption##1##2{\begingroup
    \fontsize{8.2}{9.2}\selectfont\vitalmainmakecaption{##1}{##2}\endgroup}%
  \setlength{\abovecaptionskip}{3pt}\setlength{\belowcaptionskip}{2pt}%
  \setcounter{table}{0}\setcounter{figure}{0}%
  \setcounter{topnumber}{5}\setcounter{bottomnumber}{3}\setcounter{totalnumber}{6}%
  \renewcommand{\topfraction}{0.95}\renewcommand{\bottomfraction}{0.85}%
  \renewcommand{\textfraction}{0.03}\renewcommand{\floatpagefraction}{0.80}%
  \setlength{\textfloatsep}{10pt plus 2pt minus 2pt}%
  \setlength{\floatsep}{8pt plus 2pt minus 2pt}%
  \setlength{\intextsep}{8pt plus 2pt minus 2pt}%
  \setlength{\@fptop}{0pt}%
  \setlength{\@fpsep}{8pt plus 2pt minus 2pt}%
  \setlength{\@fpbot}{0pt plus 1fil}%
  \renewcommand{\thetable}{A\arabic{table}}%
  \renewcommand{\thefigure}{A\arabic{figure}}%
  \renewcommand{\theHtable}{appendix.\arabic{table}}%
  \renewcommand{\theHfigure}{appendix.\arabic{figure}}%
}
\makeatother

\reportsetup{accent=005AFF,secondary=005AFF,resource-layout=stacked,
  title-size=23pt,title-leading=27pt,author-size=11pt,author-leading=14pt}
\renewcommand{\vitalrow}{\rowcolor{reportAbstractTint}}
\definecolor{vitalTitlePurple}{HTML}{8054BF}
\newsavebox{\vitalTitleGlyphs}
\newsavebox{\vitalTitleShading}
\DeclareRobustCommand{\vitalGradientText}[1]{%
  \begingroup
  \sbox{\vitalTitleGlyphs}{#1}%
  \leavevmode
  \sbox{\vitalTitleShading}{%
    \pdfliteral page{q 7 Tr}%
    \rlap{\usebox{\vitalTitleGlyphs}}%
    \pdfliteral page{0 Tr}%
    \tikz[baseline=0pt,inner sep=0pt,outer sep=0pt]{%
      \path[use as bounding box]
        (0,{-\dp\vitalTitleGlyphs}) rectangle
        ({\wd\vitalTitleGlyphs},{\ht\vitalTitleGlyphs});
      \shade[left color=reportAccent,right color=vitalTitlePurple]
        (0,{-\dp\vitalTitleGlyphs}) rectangle
        ({\wd\vitalTitleGlyphs},{\ht\vitalTitleGlyphs});
    }%
    \pdfliteral page{Q}%
  }%
  \pdfxform\vitalTitleShading
  \pdfrefxform\pdflastxform
  \endgroup
}
\renewcommand{\method}{\texorpdfstring{%
  {\sffamily\bfseries\upshape\vitalGradientText{VITAL}}}{VITAL}}
\title{\texorpdfstring{%
  {\fontsize{28pt}{27pt}\selectfont\vitalGradientText{VITAL}}: Visually Grounded Action Tokens for One-Step Driving Planning%
}{VITAL: Visually Grounded Action Tokens for One-Step Driving Planning}}
\reportauthor[1,2]{Alan~Liang}
\reportauthor[2,3]{Youquan~Liu}
\reportauthor[1]{Mingxiao~Li}
\reportauthor[4]{Wei~Chow}
\reportauthor[4]{Lingdong~Kong}
\reportauthorbreak
\reportauthor[4]{Yue~Liao}
\reportauthor[4]{Dongyue~Lu}
\reportauthor[5]{Yu~Yang}
\reportauthor[1]{Wei~Yin}
\reportauthor[2]{Runnan~Chen}

\makeatletter
\renewcommand\reportaffiliation[2][]{%
  \ifdefempty{\nr@affiliations}{}{%
    \ifstrequal{#1}{4}{\gappto\nr@affiliations{\par}}%
      {\gappto\nr@affiliations{,~~}}}%
  \gappto\nr@affiliations{{\normalsize\ifstrempty{#1}{}{\textsuperscript{#1}}#2}}}
\makeatother
\reportaffiliation[1]{Horizon Robotics}
\reportaffiliation[2]{Shanghai Innovation Institute}
\reportaffiliation[3]{Fudan University}
\reportaffiliation[4]{National University of Singapore}
\reportaffiliation[5]{Zhejiang University}
\reportresource[project]{\textcolor{reportAccent}{\textbf{Project Page:}}\hspace{0.4em}\ \href{https://worldbench.github.io/vital}{\texttt{https://worldbench.github.io/vital}}}{}
\reportresource[code]{\textcolor{reportAccent}{\textbf{GitHub:}}\hspace{0.4em}\ \href{https://github.com/worldbench/VITAL}{\texttt{https://github.com/worldbench/VITAL}}}{}
\newcommand{\vitalInstitutionLogos}{%
  \makebox[\linewidth][l]{%
    \includegraphics[height=11mm,trim=69bp 112bp 69bp 113bp,clip]{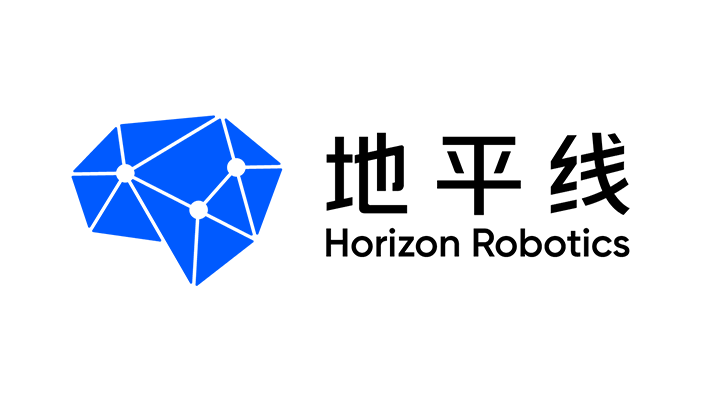}%
  }%
}

\makeatletter
\patchcmd{\nr@maketitle}
  {\vspace{6pt}{\fontsize{\nr@authorsize}{\nr@authorleading}\selectfont\nr@authors\par}}
  {\vspace{13pt}{\fontsize{\nr@authorsize}{\nr@authorleading}\selectfont\nr@authors\par}}
  {}{\PackageError{vital}{Cannot adjust title-to-author spacing}{Check the report class.}}
\patchcmd{\nr@maketitle}
  {\vspace{5pt}\nr@affiliations}
  {\vspace{7pt}\nr@affiliations}
  {}{\PackageError{vital}{Cannot adjust author-to-affiliation spacing}{Check the report class.}}
\patchcmd{\nr@maketitle}
  {\vspace{5pt}\reportresources}
  {\vspace{11pt}\reportresources}
  {}{\PackageError{vital}{Cannot adjust affiliation-to-resource spacing}{Check the report class.}}
\patchcmd{\nr@maketitle}
  {\vspace{3pt}\nr@roles}
  {\vspace{10pt}\nr@roles}
  {}{\PackageError{vital}{Cannot adjust resource-to-note spacing}{Check the report class.}}
\preto\nr@maketitle{%
  \begingroup\setlength{\parskip}{0pt}%
  \noindent\vitalInstitutionLogos\par\endgroup\vspace{-4pt}}
\renewcommand\nr@resourceformat[3]{{%
  \sffamily\fontsize{\nr@authorsize}{\nr@authorleading}\selectfont
  \makebox[1.2em][c]{\textcolor{reportAccent}{#1}}\hspace{0.4em}%
  \ifstrempty{#3}{#2}{\href{#3}{#2}}}}
\renewcommand\nr@resourceseparator{\par\vspace{5pt}}
\makeatother
\AtBeginDocument{\hypersetup{
  citecolor=reportAccent,linkcolor=reportAccent,
  pdfsubject={Vision-Induced Tokens for Action Learning},
  pdfkeywords={Autonomous driving, world-action models, action tokenization, flow matching, residual reinforcement learning},
  bookmarksnumbered=true}}

\setcitestyle{authoryear,round,semicolon}
\renewcommand{\topfraction}{.94}
\renewcommand{\bottomfraction}{.85}
\renewcommand{\textfraction}{.04}
\renewcommand{\floatpagefraction}{.80}
\renewcommand{\vitaltablefont}{\fontsize{8.2}{9.5}\selectfont}

\DeclareCaptionFont{vitalCaption}{\fontsize{11pt}{13pt}\selectfont}
\makeatletter
\AtEndCaption{\ifdefstring{\@captype}{table}{\vspace{-0.1cm}}{}}
\makeatother

\pretocmd{\section}{\Needspace{6\baselineskip}}{}%
  {\PackageError{vital}{Cannot reserve space before sections}{Check the report class.}}

\newsavebox{\vitalpagebox}
\makeatletter
\renewcommand\ps@nusreport{%
  \def\@oddfoot{%
    \begingroup
    \sffamily\bfseries
    \fontsize{\nr@pagesize}{\nr@pageleading}\selectfont
    \color{reportAccent}%
    \sbox{\vitalpagebox}{\thepage}%
    \ifnum\value{page}>1\relax
      \raisebox{\dimexpr(\ht\vitalpagebox-\dp\vitalpagebox-2.5pt)/2\relax}{%
        \rule{12mm}{2.5pt}}%
    \fi
    \hfil\usebox{\vitalpagebox}%
    \endgroup}%
  \let\@evenfoot\@oddfoot
  \let\@oddhead\@empty
  \let\@evenhead\@empty}
\patchcmd{\reportabstract}{\nr@abstract\par}{%
  \fontsize{11pt}{13.5pt}\selectfont\nr@abstract\par
}{}{\PackageError{vital}{Cannot set abstract body size}{Check the report abstract layout.}}
\makeatother

\graphicspath{{figures/}}
\abstract{\leavevmode World-action models aim to turn visual anticipation into driving decisions, yet typically train action experts on trajectory representations defined independently of visual dynamics.
The model must then align two separately constructed targets and map rich visual conditions onto sparse trajectory supervision.
We first examine a simpler setting in which trajectories are reconstructed from recorded future images.
Even without the need for future prediction, reconstruction captures the dominant maneuver but leaves centimeter-scale errors.
We therefore introduce \method{} (Vision-Induced Tokens for Action Learning), an action tokenizer that couples visual motion tokens, which decode into a trajectory scaffold, with a compact \rev{metric residual} encoding the trajectory error left by that scaffold.
Video and action supervision thus originate from the same recorded motion, while small metric corrections receive a separate, normalized learning scale.
Using flow matching, the action expert jointly predicts both components from the current image, ego state, and navigation command, without generating future images at inference.
As expected from vision-induced targets, \method{} scores 90.2 and 90.3 PDMS on NAVSIM v1 with one and ten denoising steps before any post-training.
A lightweight actor then refines execution in the residual coordinates.
With only one action denoising step, \method{} achieves 90.8 PDMS on NAVSIM v1 \texttt{navtest} at 149\,ms, $3.29\times$ the end-to-end speed of its ten-step setting.
Without retraining, it achieves 91.1 EPDMS on NAVSIM v2 \texttt{navtest} and 38.6 on \texttt{navhard}, and transfers zero-shot to nuScenes and Bench2Drive.
}

\begin{document}
\newgeometry{top=1.0cm,bottom=1.0cm,left=2.5cm,right=2.5cm}
\maketitle
\DeclareRobustCommand{\vitalteasericon}[1]{%
  \raisebox{-0.3ex}{\includegraphics[height=1.7ex]{figures/icons/#1.png}}}
\par\noindent\begin{minipage}{\linewidth}
  \setlength{\abovecaptionskip}{3pt}\setlength{\belowcaptionskip}{2pt}
  \captionsetup{font=small}
  \centering
  \includegraphics[width=\linewidth]{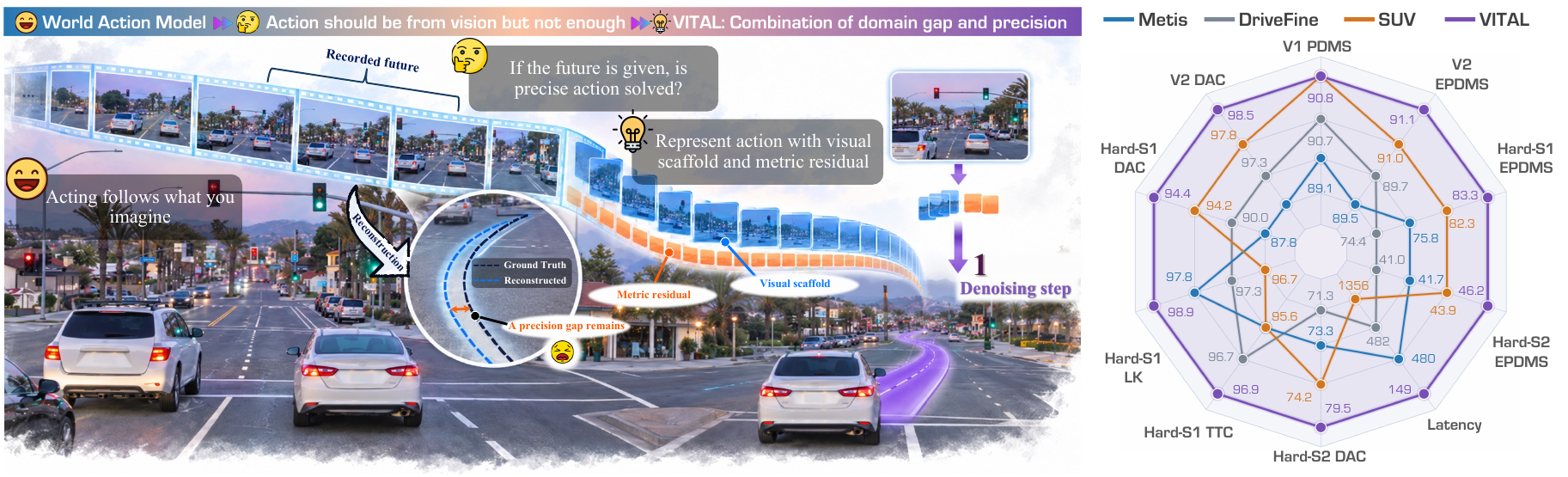}
  \vitalfigurecaption{\textbf{From world prediction to precise action.}
  \textbf{Left.} \vitalteasericon{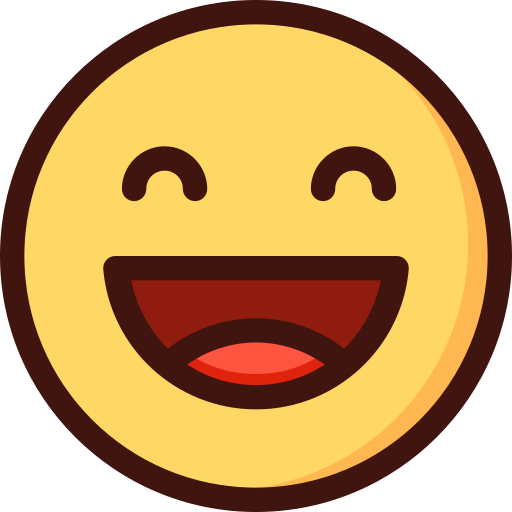}~World-action models aim to turn visual anticipation
  into executable actions. \vitalteasericon{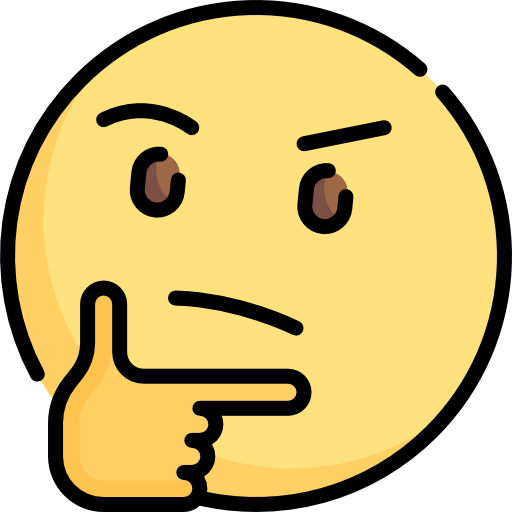}~Even with recorded future images, visual
  reconstruction leaves metric errors. \vitalteasericon{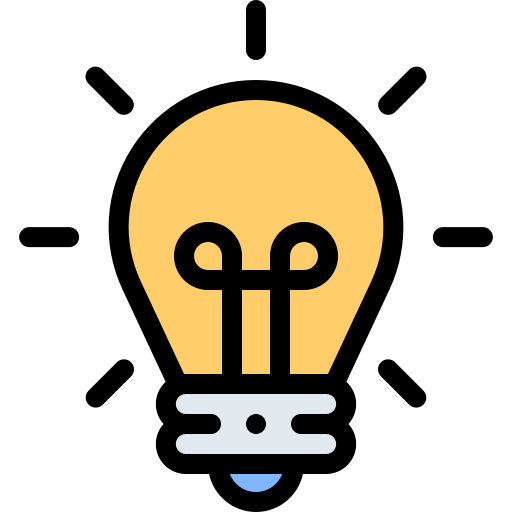}~\method{} uses visual motion as the
  principal action target and a metric residual to recover execution precision.
  \textbf{Right.} \method{} achieves the best overall performance, combining high planning quality
  across NAVSIM v1, v2, and reactive \texttt{navhard} with low inference latency compared with
  SUV~\citep{yuan2026suv}, Metis~\citep{li2026metis}, and DriveFine~\citep{dang2026drivefine}.}
  \label{fig:vital_teaser}
\end{minipage}\par\addvspace{4pt}

\noindent\begin{minipage}{\linewidth}
  \reportabstract
\end{minipage}\par
\restoregeometry
\vspace{0.2cm}
\section{Introduction}
\label{sec:introduction}

\begin{revblock}
World-action models aim to turn visual anticipation into driving decisions.
They pair a video expert, which forecasts how the scene evolves, with an action expert, which
generates the ego trajectory, and couple the two through
attention~\citep{xia2026drivelaw,zhao2026simwam,li2026metis,yuan2026suv}.
For this coupling to benefit planning, predicted scene motion must inform the trajectory, which
in turn must stay accurate to within centimeters near road boundaries and other vehicles,
as illustrated in Fig.~\ref{fig:vital_teaser}.

In current designs, this link is weakest at the action target, the representation that the
action expert learns to generate.
The video expert is trained to predict future frames, whereas the action expert typically
generates waypoints or a code constructed from trajectories alone~\citep{pertsch2025fast}, so the
two experts describe the same physical motion through separately constructed targets, and
attention must infer their correspondence.
Such targets also carry limited information about the scene.
Smooth, low-dimensional waypoints are largely determined by the ego state and navigation
command, and open-loop planners are known to exploit this shortcut at the expense of
perception~\citep{zhai2023rethinking,li2024ego}.
The action expert thus learns a target that is decoupled from the visual prediction and weakly
tied to the observed scene.

\begin{wrapfigure}{r}{.5\textwidth}
  \centering
  \captionsetup{font=small,skip=3pt}
  \includegraphics[width=\linewidth]{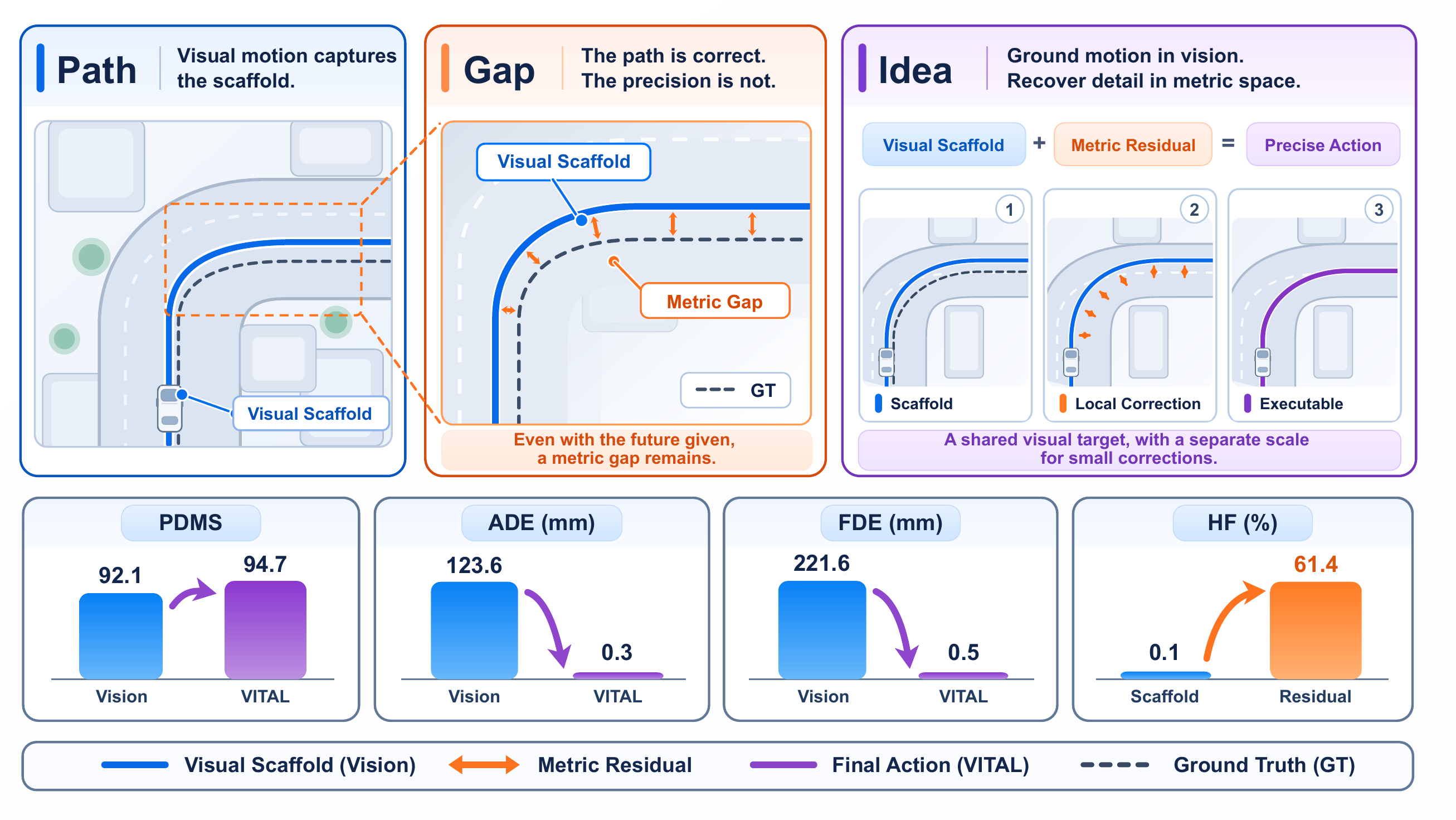}
  \caption{\textbf{Visual motion and metric precision.}
  Top, the visual scaffold captures the path and residuals restore detail.
  Bottom, reconstruction quality and high-frequency (HF) velocity energy of the scaffold and residual.}
  \label{fig:vital_visual_precision}
\end{wrapfigure}

We address both issues by deriving the action target from visual motion.
\method{} encodes the recorded frames of each training clip into \textbf{visual motion tokens}
that decode into a trajectory scaffold.
The principal action target thus originates from the same recorded motion that the video expert
learns to predict, and its dependence on the observed scene makes it difficult to infer from the
ego state and command alone.
Future frames are used only to construct targets during training, and at deployment a
flow-matching action expert generates the tokens from the current image, ego state, and
navigation command.
We therefore expect visually grounded targets to ease the visual-to-action mapping, allowing a
single flow step to produce an accurate plan.

To test whether the entire action target can be derived from vision, we first reconstruct
trajectories from recorded current and future images, removing future prediction from the task.
Surprisingly, even with these recorded futures, the visual scaffold leaves 123.6\,mm ADE and
221.6\,mm FDE, with a reconstruction PDMS of 92.1, as shown in Fig.~\ref{fig:vital_visual_precision}.
The overall driving path is recovered, but local deviations remain near road boundaries.
Accurate anticipation must therefore be accompanied by a representation that preserves the
metric detail needed to execute a driving maneuver.

We therefore complement the visual motion tokens with a compact \textbf{metric residual}
encoding the trajectory error left by the scaffold.
Adding its paired ground-truth code reduces reconstruction ADE/FDE to 0.31/0.46\,mm and
raises PDMS to 94.7.
The residual has only 1.43\% of the scaffold's XY RMS, yet its high-frequency velocity energy
fraction is 61.4\%, versus 0.1\% for the scaffold.
This small correction thus carries fine motion detail that the dominant visual path misses.
Normalizing its PCA coefficients gives these variations a learning scale of their own while
retaining visual motion as the principal action target.
A lightweight actor can later refine this code without extra denoising.
\end{revblock}

\begin{revblock}
In line with this expectation, before residual post-training \method{} scores 90.2 PDMS on NAVSIM
v1 with one denoising step and 90.3 with ten, whereas SimWAM~\citep{zhao2026simwam}, a strong WAM baseline using raw
trajectory targets, falls from 90.1 to 66.6 when limited to one step.
With residual refinement, our one-step policy reaches 90.8 PDMS on v1, 91.1 EPDMS on v2, and
38.6 EPDMS on reactive \texttt{navhard}.
As shown in the radar comparison in Fig.~\ref{fig:vital_teaser}, these results combine competitive
planning quality with 149\,ms latency per plan, the lowest reported among the compared methods.
This balance shows how visually grounded action targets translate one-step generation into
efficient planning across standard and reactive evaluations.
The representation also supports zero-shot transfer to nuScenes, Bench2Drive, and PhysicalAI
without target-dataset fine-tuning.
Our contributions are threefold.
\begin{itemize}
  \item We propose \method{}, an action tokenizer that derives a \textbf{visually grounded action
        target} from visual motion and adds a metric residual for centimeter-level precision.
  \item Through visual probes, controlled action learning, and one-step planning, we show that
        this target \textbf{eases the visual-to-action mapping}, allowing a single flow step to
        approach ten-step performance on NAVSIM benchmarks.
  \item Our model achieves \textbf{state-of-the-art one-step planning} on NAVSIM v1, v2, and
        \texttt{navhard} with lightweight residual refinement, and transfers zero-shot to three
        additional driving datasets spanning real logs and simulation.
\end{itemize}
\end{revblock}

\section{Related Work}
\label{sec:related_work}

\subsection{World-action models}

The survey of 3D and 4D world modeling organizes video-, occupancy-, and LiDAR-based approaches
and their evaluation protocols~\citep{kong2025worldsurvey}. Quo Vadis, World Modeling? broadens
this view to agent-centric proxies that provide dynamics, spatial, execution, memory, skill,
and reward feedback~\citep{yang2026quovadis}. Together, these perspectives distinguish scene
prediction from the information a policy needs to select and improve executable actions.

World-action models connect visual prediction to control by using generated futures or their latent representations to inform action generation.
Drive-WM evaluates candidate plans through generated multiview videos~\citep{wang2024driving}, while MaskGWM combines video generation with masked reconstruction to learn generalizable driving dynamics~\citep{ni2025maskgwm}.
DriveLaW injects video-generator latents into a trajectory planner~\citep{xia2026drivelaw}, and SUV lets the action expert attend to predicted appearance, semantics, relative depth, and instance dynamics~\citep{yuan2026suv}.
Metis and SimWAM use attention masks to support joint video-action training while avoiding future-video generation during action inference~\citep{li2026metis,zhao2026simwam}.

Latent world models also support driving without reconstructing full future videos.
LAW predicts action-conditioned future features to improve scene representations~\citep{li2025law}, while World4Drive incorporates intention-aware physical latent modeling~\citep{zheng2025world4drive}.
Latent-WAM combines geometry-informed scene compression with dynamic latent prediction~\citep{wang2026latentwam}.
A separate use of future prediction is to evaluate plans: WoTE forecasts BEV states for online trajectory evaluation~\citep{li2025wote}, and DA-WAM predicts an action-specific future latent for each candidate and uses it for scoring~\citep{zhong2026dawam}.

WorldLens evaluates driving world models across generation, reconstruction, action-following,
downstream tasks, and human preference~\citep{liang2026worldlens}.
The associated workshop study, Is Your Driving World Model an All-Around Player?, examines
six driving world models across 24 evaluation dimensions within the same WorldLens
framework~\citep{kong2026allaround}. These evaluations show that visual realism alone does not
establish geometric or behavioral fidelity. This motivates assessing the
control interface separately: our recorded-future reconstruction study isolates retained action
precision, while conditional prediction and NAVSIM scores test whether that representation
supports planning from current observations.

Broader surveys place both directions within agentic world modeling and vision-language-action driving systems~\citep{chu2026agentic,hu2025vlasurvey}.
LCVN studies language-conditioned navigation through latent imagination or joint observation-action prediction, while UniWM couples future prediction with memory-augmented planning~\citep{dong2026lcvn,dong2026uniwm}.
OneVL uses language and future-video supervision to learn compact reasoning tokens, discarding auxiliary decoders during planning inference~\citep{lu2026onevl}.
Beyond video, DynamicCity generates dynamic occupancy, while LiDARCrafter and U4D model controllable or uncertainty-aware LiDAR sequences~\citep{bian2025dynamiccity,liang2026lidarcrafter,xu2026u4d}. OmniLiDAR extends diffusion-based scan generation across weather, sensor configurations, and acquisition platforms through a shared range-image representation and domain-conditioned modeling~\citep{liu2026omnilidar}. These scene-generation interfaces provide a broader context for our focus on visually grounded, metric action targets.

Related strategies appear in robotic manipulation. Prediction with Action couples visual and action prediction through joint denoising~\citep{guo2024prediction}, and UWM uses independent video and action diffusion timesteps within one transformer to support policy and dynamics modeling~\citep{zhu2025uwm}. WorldVLA studies autoregressive action-world modeling~\citep{cen2025worldvla}; VideoVLA and DreamZero leverage pretrained video generators to jointly predict actions and visual outcomes~\citep{shen2025videovla,ye2026dreamzero}. Video Prediction Policy uses predictive visual representations with inverse dynamics~\citep{hu2024video}, while DreamVLA forecasts compact dynamic, spatial, and semantic knowledge for action prediction~\citep{zhang2025dreamvla}. V-JEPA 2-AC plans through an action-conditioned predictor in visual feature space~\citep{assran2025vjepa2}. Gen2Act transfers information from generated human videos to manipulation~\citep{bharadhwaj2024gen2act}, and World Guidance predicts compact conditions derived from future observations alongside actions~\citep{su2026world}.

\method{} constructs the action target from recorded visual motion and explicitly encodes the metric trajectory error left unresolved by that representation. Future images supply training supervision and target-construction inputs; deployment predicts the paired action codes from current context without generating future images or scoring trajectory candidates.

\subsection{Representation alignment and latent actions}

REPA aligns diffusion-model features with pretrained visual representations~\citep{yu2024representation}. For driving, Drive-JEPA learns predictive video representations for planning~\citep{wang2026drive}, ReWorld supervises intermediate video and action representations through prediction, alignment, and geometric discrimination~\citep{xia2026reworld}, and RAP introduces 3D geometric supervision through rasterization~\citep{feng2026rap}. Cross-modal alignment also supports 3D perception: CLIP2Scene and Seal transfer pretrained visual semantics to point clouds~\citep{chen2023clip2scene,liu2023seal}, while LargeAD scales cross-sensor pretraining and LaserMix++ exploits multimodal consistency under limited annotation~\citep{kong2026largead,kong2025multimodal}. PointLoRA adapts point-cloud representations with low-rank updates and token selection~\citep{wang2025pointlora}; Pi3DET and 3EED examine cross-platform detection and language-grounded 3D localization~\citep{liang2025pi3det,li2025threeeed}.

Representation design also shapes the fidelity of driving perception. RangeFormer addresses projection-induced semantic and geometric distortions in range-view LiDAR segmentation~\citep{kong2023rangeformer}, while LaserMix uses spatially structured beam mixing and prediction consistency to learn from sparsely labeled scans~\citep{kong2023lasermix}. RoboBEV evaluates BEV perception under camera corruptions, sensor failures, and domain shifts~\citep{xie2025robobev}. These works concern perception accuracy, label efficiency, and robustness; our complementary question is how an action representation preserves metric detail
and remains predictable from visual conditioning.

Genie and LAPA infer latent actions from visual transitions, making videos without action annotations useful for controllable generation or policy pretraining~\citep{bruce2024genie,ye2025latent}.
Moto learns motion tokens for video pretraining and subsequently connects them to robot control~\citep{chen2025moto}.
UniVLA uses language-conditioned, task-centric latent actions to reduce task-irrelevant variation~\citep{bu2025univla}.
CoMo instead learns continuous latent motion, addressing information loss in discrete representations and the gap to continuous actions~\citep{yang2026como}.
Visual transitions can contain exogenous changes, motivating supervision that separates action-relevant motion from distractors~\citep{zhang2026latent,nikulin2025latent}.

RepWAM~\citep{wang2026repwam} and LingBot-VA 2.0~\citep{zhang2026native} learn latent actions as transitions between semantically aligned visual states, with transport and residual components for reconstructing visual changes.
RLA-WM learns residual latent actions from DINO feature differences and models their distribution through flow matching~\citep{zhang2026rlawm}.
The residual in \method{} has a different target: the metric difference between a visually decoded trajectory scaffold and the demonstrated trajectory.
Trajectory supervision grounds the scaffold in physical coordinates, and a normalized compact residual code retains its remaining reconstruction errors.
Both components are jointly predicted for planning, and the residual coordinates also support bounded refinement.

\subsection{Action tokenization and reconstruction fidelity}

FAST efficiently tokenizes labeled action sequences~\citep{pertsch2025fast}.
VQ-BeT uses hierarchical vector quantization for multimodal behavior generation~\citep{lee2024vqbet}, while VQ-VLA scales vector-quantized action tokenizers~\citep{wang2025vqvla}.
QueST learns temporal skill abstractions with a flexible quantized latent representation~\citep{mete2024quest}.
OAT structures the token sequence for compression, total decodability, and ordered prediction, allowing prefix-based decoding to trade inference cost for action fidelity~\citep{liu2026oat}.
VITA additionally learns shared discrete codes that decode into future frames and robot actions~\citep{ma2026unifying}.
For autonomous driving, LaPla combines quantized trajectory representation learning with continuous latent prediction to address the discrete-to-continuous planning interface~\citep{yao2026lapla}.

\method{} instead assigns separate targets to visually grounded motion and the metric error remaining after scaffold reconstruction.

Reconstruction fidelity is only one requirement of a useful policy interface.
BAAT aligns quantized representations of behaviorally corresponding action chunks jointly with reconstruction~\citep{dong2026baat}.
ProAct explicitly studies token predictability and robustness of decoded policy outputs beyond clean action reconstruction~\citep{chen2026proact}.
Our analysis separates clean-code reconstruction from conditional code prediction, while downstream planning scores test usefulness for control; improved reconstruction alone does not establish improved planning.

\subsection{Few-step action generation}
Diffusion Policy models multimodal action sequences through iterative denoising~\citep{chi2025diffusion}, while flow matching and rectified flow learn velocity fields that transport noise to data~\citep{lipman2022flow,liu2022flow}.

Consistency Models learn direct noise-to-data mappings~\citep{song2023consistency}, and Consistency Policy and One-Step Diffusion Policy distill iterative policies into faster samplers~\citep{prasad2024consistency,wang2024one}.
Shortcut Models condition generation on step size and use self-consistency to support different sampling budgets~\citep{frans2025shortcut}; FlowPolicy applies consistency flow matching to one-step robot control~\citep{zhang2025flowpolicy}.
DanceOPD learns from capability-specific velocity fields queried at student-induced states, and DiffusionOPSD converts reward guidance into refreshed diffusion post-training targets~\citep{zhou2026danceopd,zhou2026diffusionopsd}.
Both study on-policy target construction for generative models.
MeanFlow learns average velocity over an interval~\citep{geng2026mean}, and conditional optimal-transport couplings encourage straighter flow trajectories for few-step control~\citep{sochopoulos2025fast}.
FLUX applies rectified flow to efficient cross-embodiment navigation and follows a static-to-dynamic
curriculum with reinforcement learning in social environments~\citep{gong2026flux}.
Where this approach shapes the flow and training curriculum, our approach changes the paired
action targets while retaining the standard flow-matching objective.
Streaming Flow Policy permits action execution during integration~\citep{jiang2025streaming}.

DRIFT, the residual flow adapter for VLMs, combines a coarse base predictor with iterative flow-based refinement of a localized residual distribution~\citep{liu2026residualflow}.
Its iterative refinement differs from our joint prediction of scaffold and metric residual codes.
In driving, DiffusionDrive uses truncated denoising~\citep{liao2025diffusiondrive}, while MeanFuser combines one-step generation of multiple proposals with adaptive reconstruction~\citep{wang2026meanfuser}.
\method{} retains the standard flow-matching prediction objective and changes the action representation supplied to it.
The paired scaffold and normalized metric residual support one-step prediction of a single executable trajectory without teacher distillation or candidate selection.

\subsection{Residual policy refinement}
Reward-based refinement addresses a limitation of imitation learning: matching demonstrations does not directly optimize deployment performance.
DPPO optimizes diffusion policies through reinforcement learning~\citep{ren2025diffusion}, and ReinFlow adapts flow-matching policies with online RL~\citep{zhang2025reinflow}.
Flow Q-Learning combines flow-based behavior modeling with an expressive one-step policy optimized for value~\citep{park2025fql}.

For driving, AD-R1 learns an impartial world model using counterfactual collisions and off-road
events, then uses predicted action outcomes as an internal critic for closed-loop policy
refinement~\citep{yan2026adr1}. It emphasizes the quality of learned feedback for optimization.
Simulation-grounded video--trajectory alignment also supplies matched rendered-video supervision
for reward-optimized driving trajectories~\citep{jing2026simvta}.
Our residual post-training instead uses the NAVSIM scorer during training and freezes the base
policy and codec, confining updates to bounded metric corrections around the predicted scaffold;
deployment requires neither a learned world-model critic nor reward-based candidate search.

Residual RL learns corrections around an existing controller~\citep{johannink2019residual}, while Policy Decorator and ResiP refine imitation policies through residual control~\citep{yuan2025policy,ankile2025resip}.
DSRL keeps the diffusion policy fixed and learns to steer its latent-noise inputs~\citep{wagenmaker2025dsrl}.
\method{} also freezes the base action generator and decoder, but restricts intervention to the metric residual coordinates already defined by its tokenizer.
A fixed, bounded correction space preserves the predicted visual scaffold and provides a shared physical decoding map for imitation learning and policy improvement.
At deployment, the actor applies one deterministic correction, with no extra denoising or reward-based search.
This construction bounds the scope of local refinement; it does not guarantee that every correction improves reward or satisfies task-level safety.

\section{\texorpdfstring{\rev{Visually Grounded Action Tokens}}{Visually Grounded Action Tokens}}
\label{sec:method}

\begin{revblock}
A trajectory-only code gives the action expert no explicit link to the visual motion it should
follow.
\method{} therefore derives the principal action target from \textbf{visual motion tokens} and adds
a \textbf{metric residual} for the centimeter-level detail that the visual scaffold misses.
We formulate this target in Sec.~\ref{sec:action_language}, generate it in one step in
Sec.~\ref{sec:vital_imitation}, and refine the predicted residual code in Sec.~\ref{sec:vital_posttraining}.
\end{revblock}

\subsection{\texorpdfstring{\rev{Problem Setup}}{Problem Setup}}
\label{sec:method_overview}

\begin{figure}[t]
  \centering
  \includegraphics[width=\linewidth]{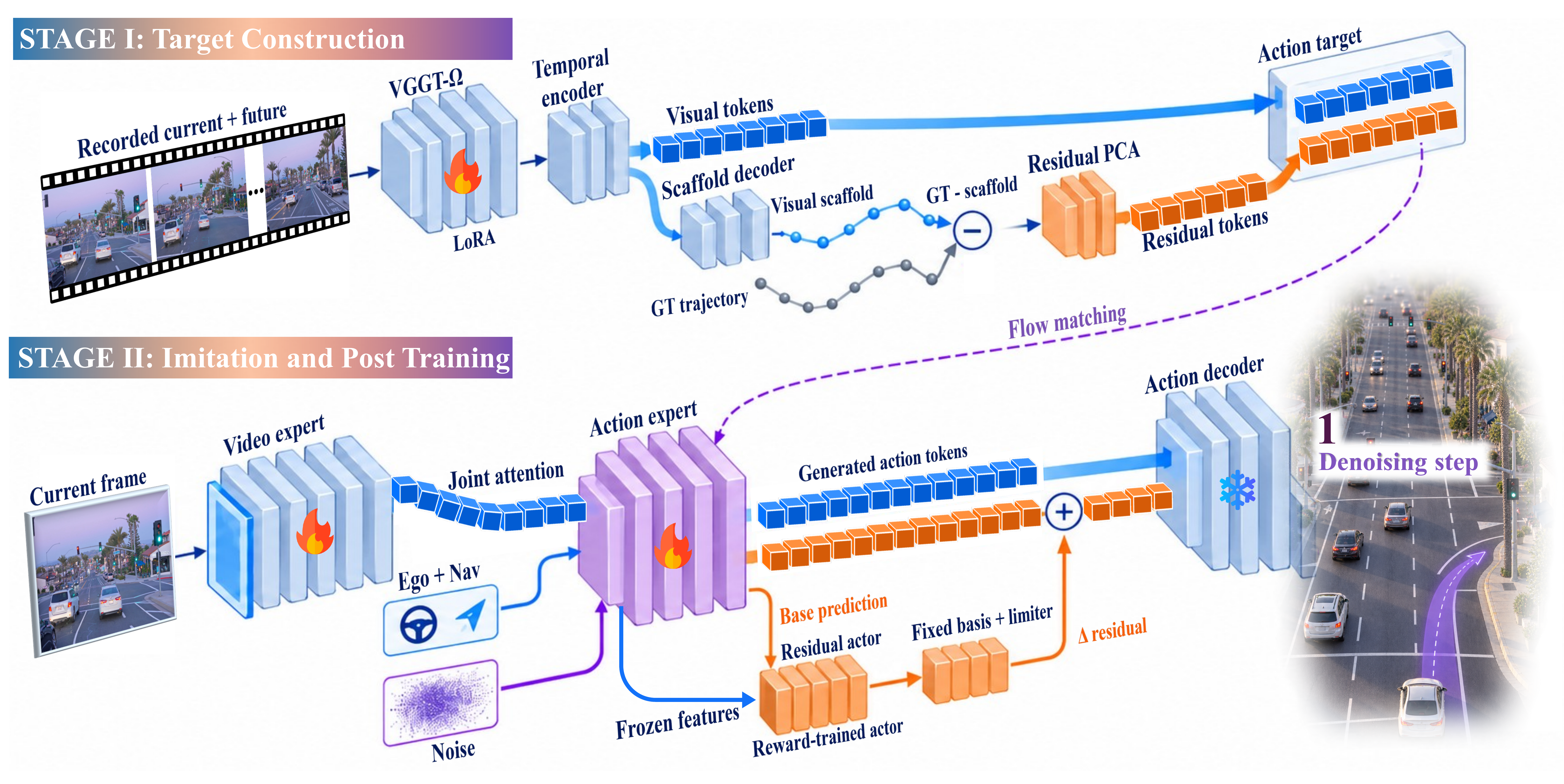}
  \caption{\rev{\textbf{Overview of \method{}.}
  In Stage I, the frozen tokenizer encodes recorded frames and the demonstrated trajectory into
  visual motion tokens, whose decoded visual scaffold represents the maneuver, and a residual
  code that captures the remaining metric error.
  In Stage II, the action expert generates both token groups from the current observation in one
  step, the residual actor applies a bounded correction, and the action decoder $D$ produces the
  executable trajectory in physical coordinates.}}
  \label{fig:vital_method}
\end{figure}

\begin{revblock}
Given a context $c$ comprising the current image, ego state, and navigation command, the planner outputs
a trajectory $\tau\in\mathbb R^{T\times d_a}$ of $T$ points with $d_a$ coordinates each.
During training, each clip also provides recorded frames $V^{+}$, consisting of the current frame
and subsequent future frames, which are used only for target construction.
Superscript stars denote targets, and hats denote predictions.
\end{revblock}
We use a video/action dual-expert WAM design, as in SimWAM~\citep{zhao2026simwam}
and Fast-WAM~\citep{yuan2026fastwam}. Action queries attend only to current-frame video tokens.
\rev{As illustrated in Fig.~\ref{fig:vital_method}, a} frozen tokenizer
maps $(V^{+},\tau)$ to visual motion tokens and a \rev{residual code}, the action expert predicts both
from $c$, and an actor refines the residual.

\subsection{\texorpdfstring{\rev{Visually Grounded Action Target}}{Visually Grounded Action Target}}
\label{sec:action_language}

\independent{\textbf{\rev{Visual Motion Tokens.}}}
\rev{To ground the action target in the motion that the video expert learns to predict, the
visual encoder $E_v$ adapts VGGT-$\Omega$~\citep{wang2026vggt} to the recorded images.}
VGGT-$\Omega$ predicts camera poses in one forward pass, and successive ego camera poses
trace the ego trajectory, making its features a natural source of motion cues.
A temporal encoder combines camera features, register summaries and correspondences, and relative
camera motion across the recorded sequence, and interval-mixing layers yield one token per temporal interval.
The scaffold decoder $D_v$ retains its camera-pose head architecture, adapting the output shape
to metric trajectory coordinates.
\begin{equation}
 z_v^\star=E_v(V^{+}),
 \qquad
 \tau_v^\star=D_v(z_v^\star).
 \label{eq:vital_scaffold}
\end{equation}
The visual motion tokens $z_v^\star\in\mathbb R^{L\times d_v}$ comprise $L$ tokens of dimension $d_v$
for successive temporal intervals, and $\tau_v^\star$ is the visual \rev{scaffold}. We fit the adaptable encoder parameters and the decoder
through trajectory reconstruction with temporal and latent regularization, then freeze the
complete visual tokenizer. The former penalizes acceleration, jerk, and interval-boundary gaps, and the latter
constrains token mean, variance, and covariance.
\rev{The tokens thus describe the maneuver in terms of the visual motion that the video expert
models, while trajectory supervision fixes their metric scale.}
\rev{Appendix~\ref{app:method_details} details the backbone, scaffold decoder, token dimensions, and fitting objectives.}

\independent{\textbf{\rev{Normalized Metric Residual.}}}
\rev{To recover metric detail without altering the visual target, we encode the scaffold's
reconstruction error $e^\star=\operatorname{vec}(\tau-\tau_v^\star)$ in a fixed linear basis.}
Residuals use a common local coordinate system for subtraction and additive decoding.
We fit a mean $\mu$, an orthonormal PCA basis $U$, and a positive diagonal coefficient scale $S$
on training residuals. These statistics define the \rev{residual code $z_r^\star$ and the action decoder $D$.}
\begin{equation}
 \begin{aligned}
 z_r^\star&=S^{-1}U^\top(e^\star-\mu),
 \qquad M=US,\\
 D(z_v,z_r)&=D_v(z_v)+\operatorname{unvec}(\mu+Mz_r).
 \end{aligned}
 \label{eq:vital_codec}
\end{equation}
The action decoder $D$ adds the decoded metric residual to the visual scaffold $D_v(z_v)$.
The \rev{residual code} has dimension $d_r$, with $U,M\in\mathbb R^{Td_a\times d_r}$ and
$S\in\mathbb R^{d_r\times d_r}$\rev{, with a fitted basis shared across all clips.}
For driving, $L=4$, $d_v=64$, and twelve residual coefficients per interval give $d_r=48$ and a
$4\times76$ \rev{action code}.
\rev{Because the basis is fitted to the errors that the scaffold actually leaves, it requires no
predefined frequency cutoff, and the scale $S$ gives centimeter-level corrections a learning scale
of their own.
The residual code enters the action expert as a second token group alongside the visual motion
tokens.}
Appendix~\ref{app:method_details} details the shared
interval basis and rank selection, and Appendix~\ref{app:method_reconstruction} measures the amplitudes
and temporal variation of the visual scaffold and metric residual.

\begin{revblock}
\independent{\textbf{Reconstruction versus Prediction.}}
\label{sec:vital_analysis}
The two components of the target address distinct sources of planning error.
A fixed trajectory norm $\|a\|_Q=(\operatorname{vec}(a)^\top Q\operatorname{vec}(a))^{1/2}$ with
$Q\succ0$ expresses all trajectory coordinates on a common physical scale.
With orthonormal $U$ and invertible $S$, $\tau$ is recovered up to the error outside the
residual subspace,
\begin{equation}
 \operatorname{vec}\!\left(\tau-D(z_v^\star,z_r^\star)\right)=(I-UU^\top)(e^\star-\mu).
 \label{eq:vital_pca_remainder}
\end{equation}
At deployment, both token groups are predicted from $c$.
Assume $D_v$ is locally $L_v$-Lipschitz from visual tokens with the Frobenius norm into the
trajectory norm.
Writing $\varepsilon_{\mathrm{codec}}=\|\tau-D(z_v^\star,z_r^\star)\|_Q$ for the reconstruction
error, $\varepsilon_v=L_v\|\hat z_v-z_v^\star\|_F$ for the visual prediction error with $L_v$ the
local Lipschitz constant of $D_v$, and $\varepsilon_r=\|M\|_{2\to Q}\|\hat z_r-z_r^\star\|_2$ for
the residual prediction error, the resulting planning error is bounded by
\begin{equation}
 \|\tau-D(\hat z_v,\hat z_r)\|_Q\le\varepsilon_{\mathrm{codec}}+\varepsilon_v+\varepsilon_r.
 \label{eq:vital_error}
\end{equation}
The bound separates representation capacity from token prediction, and $\|M\|_{2\to Q}$
converts residual code errors into trajectory errors.
A target may thus be reconstructed accurately yet remain hard to predict from $c$, so we evaluate
reconstruction, predictability, and planning separately.
The metric residual reduces $\varepsilon_{\mathrm{codec}}$, whereas the visual motion tokens are
designed to reduce $\varepsilon_v$.
Sec.~\ref{sec:ablation_results} evaluates these terms through reconstruction and conditional prediction,
while Appendix~\ref{app:method_analysis} derives the bound and the physical metric $M^\top QM$ for residual refinement.
\end{revblock}

\begin{table}[t]
\centering
\caption{\textbf{NAVSIM v1 \texttt{navtest}.} C/L denote camera/LiDAR. NFE/plans counts generator evaluations and candidates. $\dagger$ marks checkpoints evaluated by us. Other scores are cited. The visual-only planner omits the residual code and actor. Across comparison tables, \vitalbest{bold} and \vitalsecond{underlined} scores mark the best and second best per metric and setting, including ties.}
\label{tab:vital_benchmark_navtest}
\begingroup
\fontsize{6}{7.3}\selectfont
\setlength{\tabcolsep}{.55pt}
\renewcommand{\arraystretch}{1.04}
\begin{tabularx}{\linewidth}{@{}L{2.6} C{.75} C{.85} C{1.15} *{5}{C{.75}} C{.9}@{}}
\toprule
Method & Input & Frames & NFE/plans & NC\vitalup & DAC\vitalup & EP\vitalup & TTC\vitalup & C\vitalup & \textbf{PDMS}\vitalup \\
\midrule
\rowcolor{VitalLight}\multicolumn{10}{@{}l}{\textit{Regression / selection}}\\
TransFuser~\vitaltablecitep{chitta2022transfuser} & 3C+L & 1 & -- / 1 & 97.7 & 92.8 & 79.2 & 92.8 & \vitalbest{100.0} & 84.0 \\
Hydra-MDP++~\vitaltablecitep{li2025hydra} & 3C & 2 & -- / many & 97.6 & 96.0 & 80.4 & 93.1 & \vitalbest{100.0} & 86.6 \\
DiffusionDrive~\vitaltablecitep{liao2025diffusiondrive} & 3C+L & 1 & 2 / 20 & 98.2 & 96.2 & 82.2 & 94.7 & \vitalbest{100.0} & 88.1 \\
DiffusionDriveV2~\vitaltablecitep{zou2025diffusiondrivev2} & 3C+L & 1 & 2 / 20 & 98.3 & 97.9 & \vitalbest{87.5} & 94.8 & \vitalsecond{99.9} & \vitalsecond{91.2} \\
\rowcolor{VitalLight}\multicolumn{10}{@{}l}{\textit{Vision--language}}\\
ReCogDrive~\vitaltablecitep{xiong2026recogdrive} & 1C & 1 & 5 / 1 & 97.9 & 97.3 & \vitalsecond{87.3} & 94.9 & \vitalbest{100.0} & 90.8 \\
SGDrive~\vitaltablecitep{li2026sgdrive} & 1C & 4 & 5 / 1 & 98.6 & 97.8 & 85.8 & 96.2 & \vitalbest{100.0} & 91.1 \\
\rowcolor{VitalLight}\multicolumn{10}{@{}l}{\textit{Predictive / world-action}}\\
DriveLaW~\vitaltablecitep{xia2026drivelaw} & 1C & 4 & -- / 1 & \vitalsecond{99.0} & 97.1 & 81.3 & 96.7 & \vitalbest{100.0} & 89.1 \\
ReWorld~\vitaltablecitep{xia2026reworld} & 1C & 4 & 5 / 1 & \vitalbest{99.1} & 98.2 & 82.0 & \vitalbest{97.7} & 99.8 & 90.4 \\
EponaV2~\vitaltablecitep{xu2026eponav2} & 1C & 4 & -- / 1 & 98.6 & 97.9 & 84.8 & 95.7 & \vitalbest{100.0} & 90.4 \\
Metis (Top-6)~\vitaltablecitep{li2026metis} & 1C & 1 & 10 / 6 & 98.5 & 97.5 & 84.0 & 95.1 & \vitalbest{100.0} & 89.7 \\
SimWAM (RL)~\vitaltablecitep{zhao2026simwam} & 1C & 1 & 10 / 1 & 98.4 & \vitalbest{98.7} & 86.4 & 95.5 & \vitalbest{100.0} & \vitalbest{91.5} \\
SUV$^{\dagger}$~\vitaltablecitep{yuan2026suv} & 1C & 4 & 1 / 1 & \vitalsecond{99.0} & 96.8 & 84.1 & 95.9 & \vitalbest{100.0} & 89.7 \\
SUV$^{\dagger}$ & 1C & 4 & 10 / 1 & \vitalbest{99.1} & 97.8 & 84.5 & \vitalsecond{96.8} & \vitalbest{100.0} & 90.9 \\
SimWAM (IL)$^{\dagger}$~\vitaltablecitep{zhao2026simwam} & 1C & 1 & 1 / 1 & 97.3 & 91.2 & 79.1 & 81.8 & 20.5 & 66.6 \\
SimWAM (IL)$^{\dagger}$ & 1C & 1 & 10 / 1 & 98.7 & 98.0 & 83.9 & 95.7 & \vitalbest{100.0} & 90.1 \\
\rowcolor{VitalLight}\multicolumn{10}{@{}l}{\textit{One-pass, multi-proposal}}\\
MeanFuser~\vitaltablecitep{wang2026meanfuser} & 3C & 1 & 1 / multi & 98.6 & 97.0 & 82.8 & 95.0 & \vitalbest{100.0} & 89.0 \\
DRIFT~\vitaltablecitep{xing2026drift} & 1C & n.r. & 1 / 48 & 98.6 & 97.6 & 84.0 & 94.6 & \vitalbest{100.0} & 89.6 \\
\rowcolor{VitalLight}\multicolumn{10}{@{}l}{\textit{\method{} with one generated trajectory}}\\
\method{} visual tokens only & 1C & 1 & 1 / 1 & 98.7 & 97.7 & 83.8 & 95.7 & \vitalbest{100.0} & 90.0 \\
\method{} w/o residual RL & 1C & 1 & 1 / 1 & 98.6 & 98.0 & 84.1 & 95.6 & \vitalbest{100.0} & 90.2 \\
\method{} w/o residual RL & 1C & 1 & 10 / 1 & 98.6 & 98.1 & 84.1 & 95.6 & \vitalbest{100.0} & 90.3 \\
\vitalrow \textcolor{VitalPurple}{\textbf{\method{}}} & 1C & 1 & 1 / 1 & 98.7 & \vitalsecond{98.5} & 84.8 & 95.8 & \vitalbest{100.0} & \textcolor{VitalPurple}{90.8} \\
\vitalrow \method{} & 1C & 1 & 10 / 1 & 98.6 & \vitalbest{98.7} & 85.0 & 95.7 & \vitalbest{100.0} & 91.0 \\

\bottomrule
\end{tabularx}
\endgroup
\end{table}

\subsection{\texorpdfstring{\rev{One-Step Generation by Joint Flow Matching}}{One-Step Generation by Joint Flow Matching}}
\label{sec:vital_imitation}

\independent{\textbf{Joint \rev{Flow Matching}.}}
\rev{The action expert generates both token groups from $h(c)$, which comprises current-frame
video features and ego/navigation embeddings.
Future-video tokens are masked from action attention, so that the action expert relies only on
the deployment context, while $V^{+}$ serves exclusively to construct targets.}
Let $z^\star=\mathcal N(\operatorname{concat}(\operatorname{vec}(z_v^\star),z_r^\star))$ with
training-fitted affine normalization $\mathcal N$. For Gaussian noise $\epsilon\sim\mathcal N(0,I)$
and $t\in[0,1]$, we minimize the weighted joint flow-matching objective.
\begin{equation}
 \begin{aligned}
 z_t&=(1-t)\epsilon+tz^\star,
 \qquad u=z^\star-\epsilon,\\
 \mathcal L_{\mathrm{FM}}^a
 &=\mathbb E\!\left[
 w(t)\|v_\theta(z_t,t,h(c))-u\|_W^2
 \right].
 \end{aligned}
 \label{eq:vital_fm}
\end{equation}
Here $w(t)$ weights time, and the fixed positive diagonal matrix $W$ balances the two token
groups, \rev{which share the parameters of the same action expert.}

\independent{\textbf{\rev{Decoded Trajectory Loss.}}}
\rev{To supervise the executed trajectory directly, we also compare decoded predictions with the
target reconstruction in physical coordinates.}
Inverse normalization and splitting of the clean estimate
$\hat z_{\mathrm{clean}}=z_t+(1-t)v_\theta(z_t,t,h(c))$ give $(\hat z_v,\hat z_r)$.
\rev{The decoded trajectory loss and imitation-learning objective are}
{\color{VitalRev}
\begin{equation}
 \begin{aligned}
 \mathcal L_{\mathrm{dec}}
 &=\ell_\tau\big(D(\hat z_v,\operatorname{sg}[\hat z_r]),D(z_v^\star,z_r^\star)\big),\\
 \mathcal L_{\mathrm{IL}}
 &=\mathcal L_{\mathrm{FM}}^a
 +\lambda_v\mathcal L_{\mathrm{FM}}^v
 +\lambda_d\mathcal L_{\mathrm{dec}}.
 \end{aligned}
 \label{eq:vital_il}
\end{equation}}
$\mathcal L_{\mathrm{FM}}^v$ retains future-video flow matching from the video expert, and $\ell_\tau$ measures trajectory error.
\rev{The stop-gradient $\operatorname{sg}$ confines the gradients of the decoded loss to the visual
motion tokens, whereas the residual code is supervised by flow matching alone.}
Only the action expert and current-frame key/value adapters are trained, and the remaining video foundation stays frozen.

\independent{\textbf{\rev{One-Step Generation.}}}
\rev{To plan with a single evaluation of the action expert, we take one Euler step from Gaussian
noise, after which $\widetilde D$ applies inverse normalization, token splitting, and trajectory decoding.}
\begin{equation}
 \hat z^{(1)}=\epsilon+v_\theta(\epsilon,0,h(c)),
 \qquad
 \tau_{\mathrm{IL}}=\widetilde D(\hat z^{(1)}).
 \label{eq:vital_one_step}
\end{equation}
Because $z_0=\epsilon$ is independent of $(z^\star,c)$, the optimal field at $t=0$ equals
$\mathbb E[z^\star\mid h(c)]-\epsilon$, so an ideal one-step update returns the conditional mean target.
One-step planning is therefore effective when $h(c)$ nearly determines the target, \rev{a property that
the visual motion tokens are designed to provide.}
\rev{Appendix~\ref{app:planner_training} details training and sampling.}

\subsection{\texorpdfstring{\rev{Bounded Residual Refinement}}{Bounded Residual Refinement}}
\label{sec:vital_posttraining}

\rev{To improve execution while preserving the visual scaffold, we restrict reward-based
refinement to the residual code, whose coordinates correspond directly to physical corrections.}
After imitation learning, we freeze the action generator,
its visual conditioning, and the complete codec. A residual actor reads the frozen policy's
hidden states, current context, and base prediction, then outputs
$q_\phi\in\mathbb R^k$\rev{ to parameterize a correction around the base plan.}

\independent{\textbf{\rev{Bounded Correction Space.}}}
\rev{To keep corrections smooth and bounded, we project}
a family of smooth trajectory perturbations into the \rev{column space of $M$}.
Regularized whitening of the decoded Gram matrix balances the physical scales
of these directions, giving a fixed matrix $B\in\mathbb R^{d_r\times k}$.
A radial limiter $\Pi_{\mathcal C}$ rescales each proposed update to satisfy task-defined
bounds on decoded motion and token magnitude. \rev{With $\Delta z_r=\Pi_{\mathcal C}(Bq_\phi)$, the
refined plan is}
{\color{VitalRev}
\begin{equation}
 \tau_\phi=D(\hat z_v,\hat z_r+\Delta z_r)=\tau_{\mathrm{IL}}+\operatorname{unvec}(M\Delta z_r).
 \label{eq:vital_rl_policy}
\end{equation}}
The update preserves the predicted visual tokens $\hat z_v$ and their decoded scaffold. The affine residual mean
is included in the base plan, so a zero correction \rev{recovers it} exactly.

\independent{\textbf{\rev{Reward Optimization.}}}
We treat each training scenario as a contextual bandit, selecting one trajectory and
receiving its task reward. The frozen base policy identifies a hard-context pool $\mathcal H$
and a disjoint set of high-scoring anchors $\mathcal A$. Hard contexts receive reward exploration,
while \rev{on anchors the actor is regularized toward zero correction, which preserves plans that
already score well.} Gaussian exploration replaces
$q_\phi$ with $q_\phi+\sigma\xi$, where $\xi\sim\mathcal N(0,I)$, producing the perturbed plan
$\tau_\phi^\xi$. The objective combines rewards from these perturbed plans with penalties on
the actor's deterministic correction for hard contexts and anchors.
\begin{equation}
 \begin{aligned}
 J(\phi)
 &=\mathbb E_{s\sim\mathcal H,\,\xi}
       [R_\lambda(s,\tau_\phi^\xi)]
 -\beta\,\mathbb E_{s\sim\mathcal H}\Omega(q_\phi(s))\\
 &\quad-\eta\,\mathbb E_{s\sim\mathcal A}\Omega(q_\phi(s)).
 \end{aligned}
 \label{eq:vital_rl_objective}
\end{equation}
Here $R_\lambda$ combines the task reward, the official non-reactive NAVSIM v1 PDM score, with penalties for regressions in selected performance criteria.
The prior $\Omega$ penalizes the decoded mean correction before limiting. Paired positive and
negative Gaussian perturbations estimate the gradient of the smoothed reward for actor updates.
The scorer and limiter require no derivatives.
Appendix~\ref{app:method_rl} details the correction basis, radial limiter, reward pools, and \rev{gradient estimator.}

At deployment, the actor applies its deterministic mean correction once, using the same basis
and normalization as during training. \rev{The actor, trained on ten-step base plans, is reused
unchanged across inference budgets and requires no reward evaluation, candidate search, or extra
denoising.}

\section{Experiments}
\label{sec:experiments}

\subsection{Experimental Setup}
\label{sec:evaluation_protocol}

\independent{\textbf{Datasets and benchmarks.}}
NAVSIM evaluates driving trajectories in simulation~\citep{dauner2024navsim,cao2025pseudo}.
We report PDMS on v1 \texttt{navtest}, corrected EPDMS on v2 \texttt{navtest}, and reactive
two-stage EPDMS on \texttt{navhard}, together with ADE and FDE in meters. All three NAVSIM evaluations use the policy trained on v1.
We additionally evaluate zero-shot transfer to nuScenes~\citep{caesar2020nuscenes} and
Bench2Drive~\citep{jia2024bench2drive} with one NAVSIM-trained imitation checkpoint, using open-loop
L2 errors and collision rates without target-dataset fine-tuning or residual RL,
and report ADE/FDE on PhysicalAI-Autonomous-Vehicles~\citep{wang2025alpamayo}.

\independent{\textbf{Training and inference.}}
To test whether visual motion targets align video and action supervision and ease the
visual-to-action mapping, we adopt SimWAM~\citep{zhao2026simwam} as a strong driving baseline.
It follows the established video/action dual-expert design of the Fast-WAM
family~\citep{yuan2026fastwam}. Its video expert is pretrained on driving videos, providing
a suitable foundation for evaluating our action targets.

We fit the visual motion tokenizer and normalized metric residual codec on NAVSIM training scenes,
then freeze them to define the planner's joint target and trajectory decoder.
The action expert starts from existing action weights and is retrained for 26k updates to
jointly predict both targets. We update the action expert and current-frame adapters while
keeping the remaining video weights frozen, then train the residual actor for 1.6k updates
with the planner frozen.
Training uses one server with eight NVIDIA H200 GPUs.
Deployment predicts one four-second trajectory from the current front image, ego state, and navigation command.
NFE counts action expert evaluations at inference, distinct from actor training updates.
Latency at batch size one on an RTX 4090 includes image encoding,
trajectory decoding, and residual correction. Appendices~\ref{app:method_details} and~\ref{app:experimental_protocol}
detail tokenizer fitting, planner training, and evaluation.

\begin{table}[t]
\centering
\caption{\textbf{NAVSIM v2 transfer.} EPDMS on corrected \texttt{navtest} (left) and reactive two-stage \texttt{navhard} (right). \method{} retains v1 weights. Full metrics appear in Appendix~\ref{app:v2_transfer}.}
\label{tab:vital_benchmark_transfer}
\label{tab:vital_benchmark_v2}
\label{tab:vital_benchmark_navhard}
\begingroup
\fontsize{6}{7.2}\selectfont
\setlength{\tabcolsep}{.35pt}
\renewcommand{\arraystretch}{1.04}
\begin{minipage}[t]{.49\linewidth}
\begin{tabularx}{\linewidth}{@{}L{1.55} C{.75} C{.7}@{}}
\toprule
Method & NFE/plans & EPDMS\vitalup \\
\midrule
\rowcolor{VitalLight}\multicolumn{3}{@{}l}{\textit{Regression / selection}}\\
DiffusionDriveV2 \vitaltablecitep{zou2025diffusiondrivev2} & 2/20 & 87.5 \\
DriveSuprim~\vitaltablecitep{yao2026drivesuprim} & --/256 & 87.1 \\
SparseDriveV2~\vitaltablecitep{sun2026sparsedrivev2} & --/200 & 90.1 \\
\rowcolor{VitalLight}\multicolumn{3}{@{}l}{\textit{Vision--language}}\\
DriveVLA-W0~\vitaltablecitep{li2026drivevla} & AR/1 & 86.9 \\
DriveWorld-VLA~\vitaltablecitep{liu2026driveworld} & n.r. & 86.8 \\
SGDrive~\vitaltablecitep{li2026sgdrive} & 5/1 & 86.2 \\
DriveFine~\vitaltablecitep{dang2026drivefine} & $13^{*}$/1 & 89.7 \\
\rowcolor{VitalLight}\multicolumn{3}{@{}l}{\textit{Predictive / world-action}}\\
EponaV2~\vitaltablecitep{xu2026eponav2} & n.r. & 88.9 \\
Metis~\vitaltablecitep{li2026metis} & 10/1 & 89.5 \\
Metis (Top-6)~\vitaltablecitep{li2026metis} & 10/6 & 90.3 \\
SUV@1~\vitaltablecitep{yuan2026suv} & 1/1 & 89.8 \\
SUV@10 & 10/1 & 91.0 \\
\rowcolor{VitalLight}\multicolumn{3}{@{}l}{\textit{\method{}}}\\
\method{} w/o RL@1 & 1/1 & 90.7 \\
\method{} w/o RL@10 & 10/1 & 90.7 \\
\vitalrow \textcolor{VitalPurple}{\textbf{\method{}@1}} & 1/1 & \vitalsecond{91.1} \\
\vitalrow \method{}@10 & 10/1 & \vitalbest{91.2} \\
\bottomrule
\end{tabularx}
\end{minipage}\hfill
\begin{minipage}[t]{.49\linewidth}
\begin{tabularx}{\linewidth}{@{}L{1.55} C{.75} C{.7}@{}}
\toprule
Method & NFE/plans & EPDMS\vitalup \\
\midrule
\rowcolor{VitalLight}\multicolumn{3}{@{}l}{\textit{Regression / selection}}\\
LTF~\vitaltablecitep{chitta2022transfuser} & --/1 & 25.1 \\
GuideFlow~\vitaltablecitep{liu2026guideflow} & $100^{*}$/256 & 27.1 \\
\rowcolor{VitalLight}\multicolumn{3}{@{}l}{\textit{Vision--language}}\\
DriveVLA-W0~\vitaltablecitep{li2026drivevla} & AR/1 & 24.4 \\
SGDrive~\vitaltablecitep{li2026sgdrive} & 5/1 & 25.5 \\
ReCogDrive~\vitaltablecitep{xiong2026recogdrive} & 5/1 & 25.7 \\
\rowcolor{VitalLight}\multicolumn{3}{@{}l}{\textit{Predictive / world-action}}\\
EponaV2~\vitaltablecitep{xu2026eponav2} & n.r. & 36.1 \\
Metis~\vitaltablecitep{li2026metis} & 10/1 & 32.2 \\
SUV@1~\vitaltablecitep{yuan2026suv} & 1/1 & 33.0 \\
SUV@10 & 10/1 & 36.9 \\
SimWAM@10~\vitaltablecitep{zhao2026simwam} & 10/1 & 37.6 \\
GeoWAM~\vitaltablecitep{lu2026geowam} & --/1 & 36.6 \\
PerceptDrive~\vitaltablecitep{liu2026perceptdrive} & 25/1 & 34.5 \\
\rowcolor{VitalLight}\multicolumn{3}{@{}l}{\textit{\method{}}}\\
\method{} w/o RL@1 & 1/1 & \vitalsecond{38.5} \\
\method{} w/o RL@10 & 10/1 & \vitalbest{38.6} \\
\vitalrow \textcolor{VitalPurple}{\textbf{\method{}@1}} & 1/1 & \vitalbest{38.6} \\
\vitalrow \method{}@10 & 10/1 & 38.1 \\
\bottomrule
\end{tabularx}
\end{minipage}
\endgroup
\end{table}

\begin{table}[!htbp]
\centering
\caption{\textbf{Zero-shot planning on nuScenes.} Finetune marks nuScenes adaptation, and $^{*}$ denotes front camera only. \method{} uses one step without residual RL or auxiliary perception supervision.}
\label{tab:vital_benchmark_nuscenes}
\begingroup
\vitaltablefont
\setlength{\tabcolsep}{1pt}
\renewcommand{\arraystretch}{1.1}
\begin{tabularx}{\linewidth}{@{}L{2.55} C{1} C{.95} C{2} *{3}{C{.65}} C{.8} *{3}{C{.65}} C{.8}@{}}
\toprule
\multirow{2}{*}{Method} & \multirow{2}{*}{Finetune} & \multirow{2}{*}{Input} & \multirow{2}{*}{\shortstack{Auxiliary\\Supervision}} & \multicolumn{4}{c}{L2 (m)\vitaldown} & \multicolumn{4}{c}{Collision (\%)\vitaldown} \\
\cmidrule(lr){5-8}\cmidrule(lr){9-12}
& & & & 1\,s & 2\,s & 3\,s & Avg. & 1\,s & 2\,s & 3\,s & Avg. \\
\midrule
ST-P3~\vitaltablecitep{hu2022stp3} & \ding{51} & Camera & Map\&Box\&Depth & 1.33 & 2.11 & 2.90 & 2.11 & 0.23 & 0.62 & 1.27 & 0.71 \\
UniAD~\vitaltablecitep{hu2023uniad} & \ding{51} & Camera & Map\&Box\&Motion & 0.48 & 0.96 & 1.65 & 1.03 & 0.05 & 0.17 & 0.71 & 0.31 \\
VAD-Base~\vitaltablecitep{jiang2023vad} & \ding{51} & Camera & Map\&Box\&Motion & 0.54 & 1.15 & 1.98 & 1.22 & 0.04 & 0.39 & 1.17 & 0.53 \\
GenAD~\vitaltablecitep{zheng2024genad} & \ding{51} & Camera & Map\&Box\&Motion & 0.36 & 0.83 & \vitalsecond{1.55} & \vitalsecond{0.91} & 0.06 & 0.23 & 1.00 & 0.43 \\
Epona~\vitaltablecitep{zhang2025epona} & \ding{51} & Camera$^{*}$ & None & 0.61 & 1.17 & 1.98 & 1.25 & \vitalsecond{0.01} & 0.22 & 0.85 & 0.36 \\
\midrule
DriveVA~\vitaltablecitep{liu2026driveva} & \ding{55} & Camera$^{*}$ & None & 0.33 & \vitalbest{0.76} & \vitalbest{1.43} & \vitalbest{0.84} & \vitalbest{0.00} & 0.07 & \vitalbest{0.12} & \vitalbest{0.06} \\
DriveWAM~\vitaltablecitep{shi2026drivewam} & \ding{55} & Camera$^{*}$ & None & \vitalbest{0.28} & \vitalsecond{0.81} & 1.80 & 0.96 & \vitalbest{0.00} & \vitalbest{0.05} & \vitalsecond{0.14} & \vitalbest{0.06} \\
\vitalrow \textbf{\method{}} & \ding{55} & Camera$^{*}$ & None & \vitalsecond{0.30} & 0.88 & 1.85 & 1.01 & \vitalbest{0.00} & \vitalsecond{0.06} & \vitalsecond{0.14} & \vitalsecond{0.07} \\
\bottomrule
\end{tabularx}
\par
\endgroup
\end{table}

\subsection{Main Benchmark Results}
\label{sec:planning_results}

\independent{\textbf{One-step capability before residual post-training.}}
As shown in Tab.~\ref{tab:vital_benchmark_navtest}, imitation-trained \method{} scores
90.2 and 90.3 PDMS with one and ten denoising steps. The released, fully trained SimWAM
imitation checkpoint~\citep{zhao2026simwam}, evaluated with the same scorer, scores 66.6 and
90.1. Its strong ten-step score shows that extensive training still leaves a large sampling
gap. Retaining SimWAM's video/action backbone,
\method{} grounds its target in visual motion and separately normalized metric detail.
Its one-step result supports an easier visual-to-action mapping before residual RL.

\begin{wrapfigure}{R}{.5\textwidth}
  \centering
  \captionsetup{font=small,skip=3pt}
  \includegraphics[width=\linewidth,trim=0 7bp 0 2bp,clip]{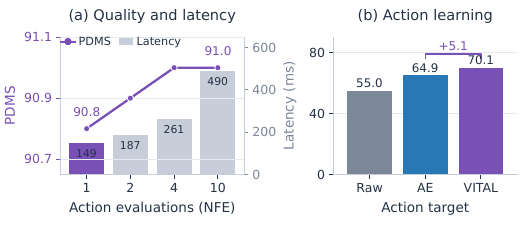}
  \caption{\textbf{Efficiency and action-target ablation.}
  (a) PDMS and median latency on v1 \texttt{navtest} across NFE.
  (b) One-step PDMS for raw trajectories, Trajectory AE, and \method{} after one training pass from a shared initialization.}
  \label{fig:vital_efficiency}
  \label{fig:vital_action_learning}
\end{wrapfigure}

\noindent\textbf{One-step planning quality and inference cost.}
As shown in Fig.~\ref{fig:vital_efficiency}a,
the final policy reaches 90.8 PDMS at 149\,ms with one step and 91.0 at 490\,ms with ten.
One step retains nearly all planning quality at $3.29\times$ the end-to-end inference speed.
Visual features are reused across evaluations, so the measured saving comes from reducing
action-flow computation. MeanFuser~\citep{wang2026meanfuser} combines multiple proposals and
DRIFT~\citep{xing2026drift} aggregates 48, while \method{} directly decodes a single generated trajectory.
The solver diagnostic in Tab.~\ref{tab:vital_solver} shows that finer integration changes
local geometry while ground-truth ADE remains nearly constant. Additional evaluations
thus improve numerical agreement without a corresponding gain in prediction accuracy.

\begin{figure}[t]
  \centering
  \includegraphics[width=\linewidth]{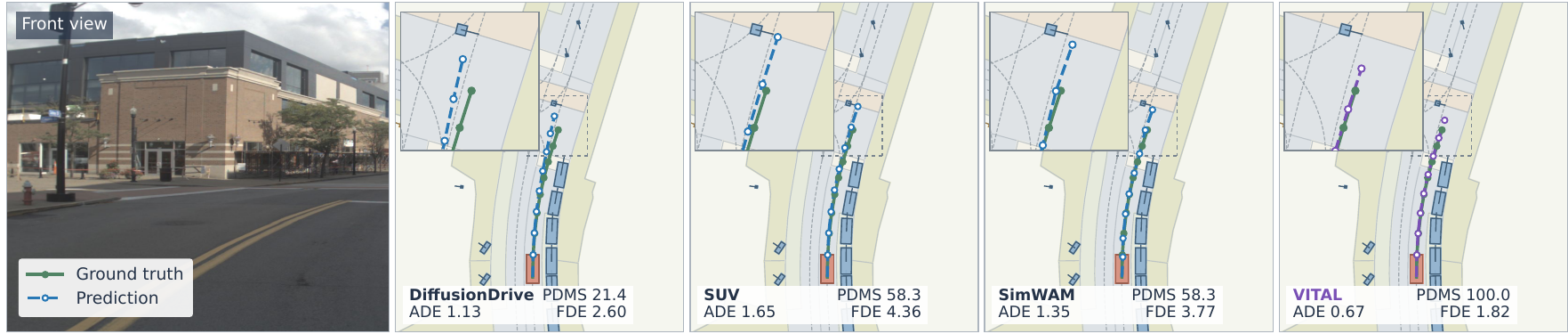}
  \caption{\textbf{Qualitative comparison on a curved road.} \method{} uses one step, SUV and SimWAM-RL ten, and DiffusionDrive two. Shared local insets show trajectory deviations near surrounding traffic. ADE and FDE are in meters.}
  \label{fig:vital_wam_comparison}
\end{figure}

\independent{\textbf{Qualitative planning comparison.}}
As shown in Fig.~\ref{fig:vital_wam_comparison}, one-step \method{} follows the bend with
0.67\,m ADE and 1.82\,m FDE, versus 1.35\,m and 3.77\,m for ten-step
SimWAM-RL~\citep{zhao2026simwam}. The shared insets show reduced endpoint overshoot near
surrounding traffic, illustrating accurate endpoint placement with one evaluation.
Appendix~\ref{app:qualitative_comparison} details the comparison protocol.

\independent{\textbf{NAVSIM v2 under corrected scoring.}}
As shown in Tab.~\ref{tab:vital_benchmark_v2}, the imitation policy reaches 90.7 EPDMS
with both one and ten evaluations under corrected v2 scoring. Its component scores also remain
closely matched across sampling budgets in Tab.~\ref{tab:vital_navtest_v2_components}, showing that the small gap extends beyond the aggregate
score. Residual post-training raises the one-step score to 91.1, compared with 91.0 for ten-step
SUV~\citep{yuan2026suv} and 90.3 for Metis with six candidates~\citep{li2026metis}.
The one-step capability persists when the same v1-trained trajectories are assessed against
the corrected driving criteria.

\independent{\textbf{NAVSIM v2 on reactive continuations.}}
As shown in Tab.~\ref{tab:vital_benchmark_navhard}, the unchanged v1 imitation policy already
reaches 38.5 and 38.6 EPDMS with one and ten evaluations on reactive two-stage \texttt{navhard}.
The final one-step policy scores 38.6, above ten-step SimWAM's 37.6~\citep{zhao2026simwam}
and its own ten-step result of 38.1. This small sampling gap supports the expectation that visually
grounded targets ease one-step generation under reactive traffic. GeoWAM forecasts future geometry from multiple cameras, and PerceptDrive
predicts future features before a 25-step flow actor~\citep{lu2026geowam,liu2026perceptdrive},
scoring 36.6 and 34.5. \method{} places visual motion in the action target and plans
from one current front image. These results support the use of visual motion targets under
both reactive traffic and synthetic observations.

\independent{\textbf{Generalization to independent datasets.}}
As shown in Tab.~\ref{tab:vital_benchmark_nuscenes}, the NAVSIM-trained imitation checkpoint
achieves 1.01\,m average L2 error and 0.07\% collision rate on nuScenes with one denoising step.
Using only the front camera, without target-dataset fine-tuning, auxiliary perception supervision,
or residual RL, it improves both averages over nuScenes-trained UniAD, VAD-Base, and Epona.
It also approaches DriveWAM's 0.96\,m and 0.06\% without nuScenes fine-tuning.
The learned visual-to-action mapping remains useful beyond its training scenes,
with low collision rates alongside accurate motion prediction.

As shown in Tab.~\ref{tab:vital_benchmark_cross_dataset}, the same checkpoint reaches 2.73\,m and
2.17\% on Bench2Drive, improving both average metrics over PWM and DriveVLA-W0. Transfer from real
logs to CARLA renderings~\citep{dosovitskiy2017carla} extends this evidence across visual domains.
Together, the two evaluations show that one-step planning persists under changes in both the
dataset and image appearance with the same learned action interface.

\begin{table}[!htbp]
\centering
\caption{\textbf{PhysicalAI planning.} ADE and FDE measure path and endpoint errors in meters at 3 and 4 seconds. \method{} uses one evaluation per plan. Dashes mark unreported four-second results.}
\label{tab:vital_benchmark_physicalai}
\begingroup
\vitaltablefont
\setlength{\tabcolsep}{.35pt}
\renewcommand{\arraystretch}{1.1}
\begin{tabularx}{\linewidth}{@{}L{2.3} *{4}{C{.675}}@{}}
\toprule
\multirow{2}{*}{Method} & \multicolumn{2}{c}{3\,s (m)\vitaldown} & \multicolumn{2}{c}{4\,s (m)\vitaldown} \\
\cmidrule(lr){2-3}\cmidrule(lr){4-5}
& ADE & FDE & ADE & FDE \\
\midrule
VaVAM~\vitaltablecitep{bartoccioni2025vavam} & 2.31 & 4.32 & -- & -- \\
SimLingo~\vitaltablecitep{renz2025simlingo} & 0.55 & 1.47 & -- & -- \\
Alpamayo-1.5~\vitaltablecitep{wang2025alpamayo} & 0.80 & 2.31 & 1.44 & 4.18 \\
DriveWAM~\vitaltablecitep{shi2026drivewam} & \vitalsecond{0.47} & \vitalsecond{1.35} & \vitalsecond{0.83} & \vitalsecond{2.47} \\
\vitalrow \textbf{\method{}} & \vitalbest{0.43} & \vitalbest{1.19} & \vitalbest{0.75} & \vitalbest{2.17} \\
\bottomrule
\end{tabularx}
\endgroup
\end{table}

\begin{table}[!htbp]
\centering
\caption{\textbf{Zero-shot transfer to nuScenes and Bench2Drive.} \method{} uses one NAVSIM imitation checkpoint on both datasets, with one step and no adaptation or residual RL. nuS. and B2D denote the two datasets. Coll. reports collision rates.}
\label{tab:vital_benchmark_cross_dataset}
\begingroup
\vitaltablefont
\setlength{\tabcolsep}{.35pt}
\renewcommand{\arraystretch}{1.1}
\begin{tabularx}{\linewidth}{@{}L{3.4} *{3}{C{.8}} C{1} @{\hspace{2pt}} *{3}{C{.8}} C{1} @{\hspace{3pt}} *{3}{C{.8}} C{1} @{\hspace{2pt}} *{3}{C{.8}} C{1}@{}}
\toprule
\multirow{2}{*}{Method} & \multicolumn{4}{c}{\textbf{nuS.} L2 (m)\vitaldown} & \multicolumn{4}{c}{\textbf{nuS.} Coll. (\%)\vitaldown} & \multicolumn{4}{c}{\textbf{B2D} L2 (m)\vitaldown} & \multicolumn{4}{c}{\textbf{B2D} Coll. (\%)\vitaldown} \\
\cmidrule(lr){2-5}\cmidrule(lr){6-9}\cmidrule(lr){10-13}\cmidrule(lr){14-17}
& 1\,s & 2\,s & 3\,s & Avg. & 1\,s & 2\,s & 3\,s & Avg. & 1\,s & 2\,s & 3\,s & Avg. & 1\,s & 2\,s & 3\,s & Avg. \\
\midrule
DriveVLA-W0~\vitaltablecitep{li2026drivevla} & 0.43 & 1.26 & 2.60 & 1.43 & 0.22 & 0.66 & 1.42 & 0.77 & \vitalsecond{1.01} & 2.77 & 5.22 & 3.00 & 1.49 & 2.53 & \vitalsecond{3.53} & 2.52 \\
PWM~\vitaltablecitep{zhao2025pwm} & 2.06 & 3.91 & 6.00 & 3.99 & \vitalsecond{0.12} & 0.15 & 0.86 & 0.36 & 1.70 & 2.74 & \vitalsecond{3.97} & 2.80 & 4.01 & 3.73 & \vitalsecond{3.53} & 3.76 \\
DriveVA~\vitaltablecitep{liu2026driveva} & \vitalsecond{0.33} & \vitalbest{0.76} & \vitalbest{1.43} & \vitalbest{0.84} & \vitalbest{0.00} & \vitalsecond{0.07} & \vitalbest{0.12} & \vitalbest{0.06} & \vitalbest{0.69} & \vitalbest{1.29} & \vitalbest{2.03} & \vitalbest{1.33} & \vitalsecond{1.38} & \vitalbest{1.97} & \vitalbest{2.65} & \vitalbest{1.79} \\
\vitalrow \textbf{\method{}} & \vitalbest{0.30} & \vitalsecond{0.88} & \vitalsecond{1.85} & \vitalsecond{1.01} & \vitalbest{0.00} & \vitalbest{0.06} & \vitalsecond{0.14} & \vitalsecond{0.07} & 1.04 & \vitalsecond{2.62} & 4.52 & \vitalsecond{2.73} & \vitalbest{0.74} & \vitalsecond{2.19} & 3.58 & \vitalsecond{2.17} \\
\bottomrule
\end{tabularx}
\par
\endgroup
\end{table}

\begin{table}[!htbp]
\centering
\caption{\textbf{Recorded-image reconstruction.}}
\label{tab:vital_codec_ablation}
\begingroup
\vitaltablefont
\setlength{\tabcolsep}{.45pt}
\renewcommand{\arraystretch}{1.1}
\begin{tabularx}{\linewidth}{@{}L{1.05} C{.98333} C{.98333} C{.98334}@{}}
\toprule
\multirow{2}{*}{\shortstack{Code\\pairing}} & \multicolumn{3}{c}{Reconstruction error\vitaldown} \\
\cmidrule(lr){2-4}
& ADE (m) & FDE (m) & Yaw ($^{\circ}$) \\
\midrule
Scaffold only & \vitalsecond{0.12} & \vitalsecond{0.22} & \vitalsecond{0.43} \\
Zero residual & \vitalsecond{0.12} & \vitalsecond{0.22} & \vitalsecond{0.43} \\
\vitalrow Matched pair & $\vitalbest{3.11\times10^{-4}}$ & $\vitalbest{4.62\times10^{-4}}$ & $\vitalbest{3.38\times10^{-3}}$ \\
Shuffle visual & 7.41 & 15.10 & 14.70 \\
Shuffle residual & 0.19 & 0.34 & 0.65 \\
\bottomrule
\end{tabularx}
\endgroup
\end{table}

\Needspace{6\baselineskip}
\independent{\textbf{Motion accuracy on PhysicalAI.}}
Tab.~\ref{tab:vital_benchmark_physicalai} reports \method{}'s ADE/FDE of
0.43/1.19\,m at 3\,s and 0.75/2.17\,m at 4\,s, the lowest errors among the listed methods.
Relative to DriveWAM, FDE decreases by 12.1\% at 3\,s and 12.0\% at 4\,s.
The endpoint gains at both horizons complement the path-level results, supporting the
representation's ability to retain metric detail during one-step planning.

\subsection{Ablation Studies}
\label{sec:ablation_results}

\Needspace{3\baselineskip}
\phantomsection
\independent{\textbf{Ablation on the Metric Residual.}}
\label{sec:codec_results}
Even with recorded current and future images, the visual scaffold leaves 123.6\,mm ADE,
reduced to 0.31\,mm by its paired ground-truth residual in Fig.~\ref{fig:vital_visual_precision}
and Tab.~\ref{tab:vital_codec_ablation}. Shuffling visual codes raises ADE to 7.41\,m, versus
0.19\,m when shuffling residual codes. These controls place the principal path in the visual
tokens and the remaining metric detail in the residual. Rank 12 retains this precision with
twelve coefficients per interval, as shown in Tab.~\ref{tab:vital_residual_rank}, and
Appendix~\ref{app:method_reconstruction} documents its finer velocity variation. For targets predicted
from $c$, the matched one-step planners in Tab.~\ref{tab:vital_benchmark_navtest} run without
the residual actor. Visual tokens alone reach 90.0 PDMS, while the predicted residual raises
PDMS to 90.2, DAC from 97.7 to 98.0, and EP from 83.8 to 84.1. Visual motion therefore carries
most of the one-step planning ability, with the compact residual improving road compliance
and progress through precise local metric corrections.

\begin{table}[t]
\centering
\begingroup
\captionsetup{font=footnotesize,justification=raggedright,singlelinecheck=false}
\setlength{\abovecaptionskip}{3pt}\setlength{\belowcaptionskip}{3pt}
\begin{minipage}[t]{.49\linewidth}
\vspace{0pt}
\begingroup
\captionsetup{belowskip=3pt}
\vitaltablecaption{\textbf{Visual probes.} Ego/nav. alone $\to$ with vision. Both targets use equal-capacity probes.}
\label{tab:vital_ablations}
\centering\fontsize{6}{7.2}\selectfont
\setlength{\tabcolsep}{.35pt}\renewcommand{\arraystretch}{1.1}
\begin{tabularx}{\linewidth}{@{}L{1.21} C{.97} C{.97} C{.85}@{}}
\toprule
Target & ADE (m)$\downarrow$ & PDMS$\uparrow$ & MSE$_{\mathrm{V}}\downarrow$ \\
\midrule
Trajectory AE & \mbox{\vitalbest{1.08}$\!\to\!$\vitalsecond{0.80}} & \mbox{\vitalbest{57.3}$\!\to\!$\vitalsecond{73.2}} & \vitalbest{0.12} \\
\method{} shuf. & \mbox{\vitalsecond{1.62}$\!\to\!$2.20} & \mbox{\vitalsecond{56.4}$\!\to\!$52.3} & -- \\
\vitalrow \method{} & \mbox{\vitalsecond{1.62}$\!\to\!$\vitalbest{0.72}} & \mbox{\vitalsecond{56.4}$\!\to\!$\vitalbest{79.4}} & \vitalsecond{0.38} \\
\bottomrule
\end{tabularx}
\par
\endgroup

\end{minipage}\hfill
\begin{minipage}[t]{.49\linewidth}
\vspace{0pt}
\begingroup
\captionsetup{belowskip=3pt}
\vitaltablecaption{\textbf{Action-target ablation.} Shared initialization and one training pass without residual RL.}
\label{tab:vital_action_learning}
\centering\fontsize{6}{7.2}\selectfont
\setlength{\tabcolsep}{.35pt}\renewcommand{\arraystretch}{1.1}
\begin{tabularx}{\linewidth}{@{}L{1.3} C{.8} C{.95} C{.95}@{}}
\toprule
Action target & PDMS$\uparrow$ & ADE (m)$\downarrow$ & FDE (m)$\downarrow$ \\
\midrule
Raw trajectory & 55.0 & 1.46 & 3.49 \\
Trajectory AE & \vitalsecond{64.9} & \vitalsecond{1.25} & \vitalsecond{2.45} \\
\vitalrow \method{} & \vitalbest{70.1} & \vitalbest{1.16} & \vitalbest{2.13} \\
\bottomrule
\end{tabularx}
\par
\endgroup

\end{minipage}
\endgroup
\end{table}

\Needspace{3\baselineskip}
\phantomsection
\independent{\textbf{Ablation on Visual Conditioning.}}
\label{sec:representation_results}
Equal-capacity ridge probes predict \method{} and Trajectory AE codes, each with 304 coordinates,
from the same frozen features. Ego state and navigation alone favor AE in
Tab.~\ref{tab:vital_ablations}, with 1.08\,m ADE versus \method{}'s 1.62\,m. Matched current-image
features reverse this ordering, giving \method{} 0.72\,m ADE and 79.4 PDMS versus AE's
0.80\,m and 73.2 PDMS. Shuffling vision drops \method{} to 52.3 PDMS. The reversal and
shuffling control show that its larger gain comes from scene correspondence, supporting visual
motion as an action target that benefits from the observed context. Its higher standardized
token MSE of 0.38 versus AE's 0.12 coexists with better decoded paths, consistent with the
decoder-dependent error geometry in Sec.~\ref{sec:vital_analysis}.
Appendix~\ref{app:representation_controls} reports the complete protocol and additional controls for both targets.

\Needspace{3\baselineskip}
\phantomsection
\independent{\textbf{Ablation on Action Targets.}}
\label{sec:target_results}
Raw trajectories, Trajectory AE, and \method{} share the pretrained visual foundation,
random action-core initialization, training batches, optimizer,
schedule, and evaluation scenes. AE matches \method{}'s $4\times76$ token shape, while raw
prediction uses SimWAM's $8\times3$ waypoints~\citep{zhao2026simwam}. We train all three
for one training pass without residual RL, testing how effectively each target converts
the same visual supervision into planning behavior. As shown in Fig.~\ref{fig:vital_action_learning}b
and Tab.~\ref{tab:vital_action_learning}, one-step \method{} reaches 70.1 PDMS versus AE's
64.9 and raw prediction's 55.0, with the lowest ADE and FDE.

AE's gain over raw prediction demonstrates the value of learned coding. With the same token
shape and training exposure, \method{} further improves road compliance and progress at both
sampling budgets, as shown in Tab.~\ref{tab:vital_action_sampling}. Together with the visual
probes, this advantage supports our expectation that visual motion targets align supervision
and ease the conditional mapping under limited updates. The fully trained SimWAM comparison
in Tab.~\ref{tab:vital_benchmark_navtest} also supports this one-step advantage after
extensive training.

\Needspace{3\baselineskip}
\independent{\textbf{Ablation on Residual RL.}}
An actor with 1.23M parameters refines six bounded metric directions with the generator and
visual scaffold frozen, adding 0.6 and 0.7 PDMS at one and ten evaluations. The three-seed
controls at NFE$=10$ in Tab.~\ref{tab:vital_rl} show positive mean gains across reward pools
and prior weights, with broader reward coverage helping more than a weaker prior. Using
non-reactive v1 rewards, the same actor raises one-step v2 EPDMS from 90.7 to 91.1.
In Tab.~\ref{tab:vital_transfer_details}, \texttt{navhard} Stage~1 improves from 82.3 to 83.3,
while Stage~2 declines from 47.1 to 46.2, leaving the combined score at 38.6 versus 38.5 before
RL. This stage contrast is consistent with the expected scope of targeted refinement.
The actor optimizes local execution around a fixed scaffold on logged v1 scenes, whereas
shifted states and reactive traffic in Stage~2 can require revising the maneuver itself.
Such adaptation calls for changes to the scaffold beyond the bounded residual corrections
learned by the actor. The IL planner already reaches 38.5 EPDMS, and the final one-step
38.6 exceeds ten-step SimWAM's 37.6 in Tab.~\ref{tab:vital_benchmark_navhard}.
The representation thus retains strong reactive planning quality while residual RL adds
targeted execution gains on both v1 and v2 \texttt{navtest}.

\section{Conclusion}
\label{sec:conclusion}

We present \method{}, a world-action planning framework that jointly addresses
visual--action alignment and metric precision. Deriving the principal action target from
recorded visual motion lets video and action supervision describe the same motion and eases the
visual-to-action mapping, while a normalized metric residual restores precision and hosts lightweight
post-training. With only one denoising step, \method{} achieves 90.8 PDMS on NAVSIM v1, 91.1 EPDMS on v2, and 38.6 EPDMS on \texttt{navhard},
and transfers zero-shot to nuScenes and Bench2Drive. Visually grounded action targets directly connect scene anticipation to planning.

\FloatBarrier
\section*{LLM Usage Statement}

We used generative AI tools solely for language polishing to improve grammar, clarity,
and readability. They were not used to develop the research ideas or methods, conduct
experiments, or analyze or interpret results. The authors reviewed all AI-assisted edits
to ensure that the scientific meaning was preserved and take full responsibility for
the final manuscript.

\clearpage
\appendix
\vitalappendixfloats
\begingroup
\setcounter{tocdepth}{3}
\makeatletter
\section*{\fontsize{20pt}{24pt}\selectfont Appendix}
Appendix~\ref{app:method_details} develops the visual scaffold, metric residual, and causal planner.
Appendix~\ref{app:experimental_protocol} tests visual conditioning, transfer across scoring protocols, and reuse of the action tokenizer across planners.
Appendices~\ref{app:qualitative} to~\ref{app:public_resources} compare trajectories, discuss the scope of local refinement, and list the public resources used.
\let\vital@savedsectionentry\l@section
\renewcommand{\l@section}[2]{%
  \begingroup\fontsize{12pt}{14.5pt}\selectfont
  \vital@savedsectionentry{#1}{#2}\endgroup}
\let\vital@savedcontentsline\contentsline
\def\vital@firstappendix{appendix.A}
\newif\ifvital@appendixentry
\vital@appendixentryfalse
\def\contentsline#1#2#3#4{%
  \def\vital@entryanchor{#4}%
  \ifx\vital@entryanchor\vital@firstappendix
    \vital@appendixentrytrue
  \fi
  \ifvital@appendixentry
    \vital@savedcontentsline{#1}{#2}{#3}{#4}%
  \fi}
\@starttoc{toc}
\makeatother
\endgroup
\clearpage

\section{Method Details and Analysis}
\label{app:method_details}

\subsection{Target Representation}
\label{app:method_reconstruction}
\independent{\textbf{From recorded motion to an action target.}}
The tokenizer pairs visual motion tokens $z_v^\star=E_v(V^+)$ with a metric residual code
$z_r^\star$ for the error left by $D_v(z_v^\star)$. The action expert predicts both targets jointly
from the current front image, ego state, and navigation command, as defined in
Section~\ref{sec:method_overview}. For a horizon of $T=40$ samples at 10\,Hz, each pose has
$d_a=3$ ego-frame coordinates. Each of the $L=4$ intervals pairs a visual token of width $d_v=64$
with $d_r/L=12$ residual coefficients. The resulting $4\times76$ tensor pairs visual motion and
the corresponding metric correction at each interval within the four-second prediction horizon.

\independent{\textbf{Visual tokenizer.}}
The target encoder $E_v$ adapts a frozen VGGT-$\Omega$~\citep{wang2026vggt} backbone through
LoRA~\citep{hu2022lora}. It takes five front images at $0,1,2,3,4$ seconds. Two temporal layers
and two interval mixing layers combine camera features, register summaries, register
correspondences, and relative camera motion. The scaffold decoder $D_v$ adapts the camera pose
head architecture to output bounded body-frame velocities and yaw rates with hidden width 256.
Integrating these rates gives the scaffold on the trajectory grid. Trajectory supervision sets the
metric scale for motion estimated from monocular images.

The tokenizer also learns to reconstruct perturbed codes and to predict codes with an auxiliary
flow objective. Its 256-dimensional conditioning feature combines the first-frame camera head
state, mean and attention pooled registers, and relative camera motion before temporal mixing.
Latent velocity and decoded one-step estimates both supervise fitting. We select the tokenizer
by held-out reconstruction error and freeze its weights before fitting the residual. The causal
planner in Appendix~\ref{app:planner_training} uses the resulting tokens as targets rather than observed inputs.

\independent{\textbf{Tokenizer objectives.}}
Write the tokenizer objective as a weighted sum $\sum_j\lambda_j\mathcal L_j$.
For position, increments, keyframes, and heading, the weights are $(1,0.25,0.75,0.1)$.
Noisy code reconstruction uses weight $0.15$ and noise scale $\sigma_z=0.10$. The latent flow
and decoded probe weights both reach $0.05$ after a 3,000 update ramp. Body velocity, yaw rate,
acceleration, jerk, and boundary continuity use weights $(0.1,0.02,0.002,2\cdot10^{-5},0.01)$,
while latent mean, variance, and covariance use $(0.01,0.02,0.05)$. Eq.~\ref{eq:vital_il} defines
the separate planner objective used after freezing the tokenizer.

\independent{\textbf{Paired metric residual coordinates.}}
We stack the $T/L$ ego-frame pose errors in interval $j$ as $e_j\in\mathbb R^{30}$, with positions
in meters and heading in radians. Angular differences $\delta$ are wrapped using
$\operatorname{atan2}(\sin\delta,\cos\delta)$. Fitting PCA to these training residuals gives a
shared mean $\mu_0$, basis $U_0$, and diagonal matrix $S_0$ of coefficient standard deviations.
Repeating the interval statistics builds the full residual decoder,
\begin{equation}
 U=I_4\otimes U_0,\quad S=I_4\otimes S_0,\quad
 M=US\in\mathbb R^{120\times48},\quad
 \mu=\mathbf 1_4\otimes\mu_0.
 \label{eq:vital_blocks}
\end{equation}
where $\mu+Mz_r$ decodes the residual in pose space. Scaling by $S^{-1}$ standardizes the residual
before concatenation. The outer normalization $\mathcal N$ then standardizes the complete joint target using fixed training statistics.

We sweep interval ranks $\{2,4,6,8,9,10,11,12,16,24,30\}$.
Tab.~\ref{tab:vital_residual_rank} shows why we use rank 12. It reaches submillimeter mean
reconstruction error while using four fewer coefficients per interval than rank 16.

\independent{\textbf{Motion scale and temporal variation.}}
\label{app:motion_scales}
On the training partition, we measure the decoded visual scaffolds and their \rev{metric residuals} on the
40-point, 10\,Hz grid. We compute orthonormal DCT-II spectra after removing each pose sequence's
temporal mean. Velocity uses the 39 forward differences $(p_{i+1}-p_i)/(0.1\,\mathrm{s})$ without
demeaning. Bins $k\in[0,4)$ define low-frequency energy, and the remaining bins define higher-frequency
energy. Tab.~\ref{tab:vital_motion_scales} pools squared coefficients over scenes. Fig.~\ref{fig:vital_visual_precision}
averages the velocity fractions over $x$ and $y$. Before mean removal the residual's XY RMS
is $1.43\%$ of the scaffold's. It adds small corrections at both time scales and carries a larger
share of high-frequency velocity energy than the scaffold.
\begin{table}[!htbp]
\centering
\caption{\textbf{Residual-rank reconstruction.} The table compares held-out mean and 99th percentile reconstruction errors in millimeters.}
\label{tab:vital_residual_rank}
\begingroup\vitaltablefont
\setlength{\tabcolsep}{1.1pt}\renewcommand{\arraystretch}{1.04}
\begin{tabularx}{.72\linewidth}{@{}L{1.9} C{0.82}C{0.82}C{0.82}C{0.82}C{0.82}@{}}
\toprule
Rank & 8 & 10 & 11 & 12 & 16 \\
\midrule
Mean error\vitaldown & 2.76 & 0.74 & 0.73 & 0.31 & 0.06 \\
P99 error\vitaldown & 10.20 & 1.48 & 1.47 & 1.02 & 0.19 \\
\bottomrule
\end{tabularx}
\endgroup
\end{table}

\begin{table}[!htbp]
\centering
\caption{\textbf{Temporal energy proportions (\%).} LF and HF denote low and high frequencies of the scaffold and its paired residual.}
\label{tab:vital_motion_scales}
\begingroup\vitaltablefont
\setlength{\tabcolsep}{1.1pt}\renewcommand{\arraystretch}{1.04}
\begin{tabularx}{.72\linewidth}{@{}L{1.9} C{0.775}C{0.775}C{0.775}C{0.775}@{}}
\toprule
& \multicolumn{2}{c}{Pose LF} & \multicolumn{2}{c}{Velocity HF} \\ \cmidrule(lr){2-3}\cmidrule(l){4-5} Component & $x$ & $y$ & $x$ & $y$ \\
\midrule
Visual scaffold & 99.70 & 99.50 & 0.05 & 0.10 \\
Conditional residual & 91.90 & 92.50 & 52.90 & 69.90 \\
\bottomrule
\end{tabularx}
\endgroup
\end{table}

\independent{\textbf{Trajectory reconstruction accuracy.}}
\label{app:visual_precision_diagnostic}
As shown in Fig.~\ref{fig:vital_visual_precision} and Tab.~\ref{tab:vital_codec_ablation}, adding the
paired recorded residual reduces reconstruction ADE from 123.60 to 0.31\,mm on the full
validation split. Both variants use the same recorded frames. This measures how much trajectory
detail the representation retains, while PDMS measures execution against the PDM reference.

Both variants share the VGGT-$\Omega$ aggregator and camera head with 1,111,483,401 base
parameters. Beyond LoRA, the visual tokenizer fits 6.19M parameters across the temporal encoder
(4,162,010), scaffold decoder (219,523), and auxiliary predictor (1,811,520).
The residual adds 402 fixed scalars for the statistics in Eq.~\ref{eq:vital_blocks}.

\begin{figure}[!htbp]
  \centering
  \includegraphics[width=.86\linewidth]{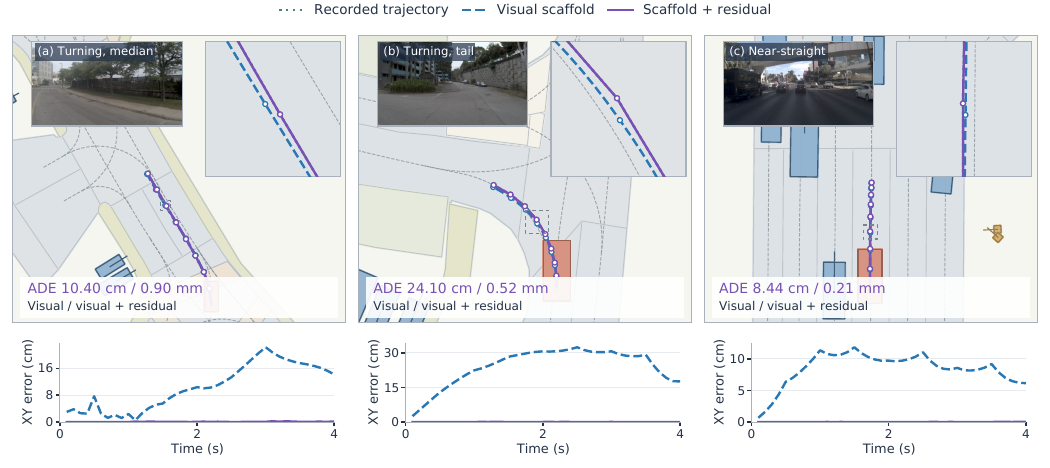}
  \caption{\textbf{Clean-code trajectory reconstruction.} The visual scaffold and paired residual together recover the recorded path.}
  \label{fig:vital_codec_cases}
\end{figure}

\independent{\textbf{Clean-code examples.}}
\label{app:qualitative_codec}
Fig.~\ref{fig:vital_codec_cases} shows how the two targets divide the reconstruction task.
Visual tokens recover the route shape, and the metric residual restores local alignment with the
recorded trajectory in both turning and near-straight scenes.

\subsection{Causal Planner Training}
\label{app:planner_training}

\independent{\textbf{Planner training and causal conditioning.}}
We initialize the video expert from SimWAM~\citep{zhao2026simwam} and use a dual expert design
with action-only inference, as in Fast-WAM~\citep{yuan2026fastwam}. A causal mask lets action
and video queries read current-frame keys and values, while future video tokens are visible
only to video queries. We train the action expert and rank-16 key and value adapters on the
current-frame prefix of the last eight video blocks. All examples share a fixed text prompt,
and inference reuses the same adapted current-frame features across all evaluations of the action flow.

The main action expert starts from weights pretrained for 74,000 action-only updates and
56,000 joint video and action updates. We then train the planner on paired \method{} targets
for 26,000 updates on eight NVIDIA H200 GPUs, with AdamW, peak learning rate
$2\times10^{-5}$, and effective batch size 64. Appendix~\ref{app:action_learning} instead
compares targets from a shared random action-core initialization and one pass through the training split, using identical batches.

\independent{\textbf{Flow matching and sampling conventions.}}
We use the data-to-noise fraction $\sigma$, with
$x_\sigma=(1-\sigma)z^\star+\sigma\epsilon$ and target velocity $\epsilon-z^\star$.
Training uses the schedule $\varphi_a(u)=au/[1+(a-1)u]$ with $a=5$ and
$u\sim\mathcal U(0,1)$, along with the scheduler's positive interior time weight.
Eq.~\ref{eq:vital_fm} uses $t=1-\sigma$ and
$v_\theta(z,t,h)=-f_\theta(z,1000(1-t),h)$.
The clean estimate and inverse-normalized token split are
\begin{equation}
 \hat z_{\mathrm{clean}}=z_t+(1-t)v_\theta(z_t,t,h(c)),
 \qquad
 (\hat z_v,\hat z_r)=\operatorname{split}\!\left(\mathcal N^{-1}(\hat z_{\mathrm{clean}})\right).
 \label{eq:vital_clean_estimate}
\end{equation}
At inference, $\varphi_5(1-k/K)$ sets the grid for $K$ evaluations of the same velocity field.
At $K=1$, the sampler advances from the initial noise to the data endpoint in one Euler update.

Visual and residual dimensions have relative flow-matching weights $1:2$, normalized to mean
one across the 76 coordinates. The visual prefix therefore accounts for
$64/(64+24)=72.7\%$ of the dimension-weight mass. The imitation objective combines action
flow matching with video flow matching weighted by $\lambda_v=0.25$. Both token groups are
learned jointly through action flow matching. The scaffold decoder and affine residual transform
both use FP32 so that decoding preserves the small position and heading changes in the predicted residual.

\subsection{Residual Policy, Geometry and Reward Optimization}
\label{app:method_rl}

\independent{\textbf{A fixed correction family.}}
We use a scene-independent basis with $k=6$ directions. For normalized time $\rho_i=i/T$,
the templates perturb longitudinal and lateral positions by $\rho_i^p$ for $p\in\{2,3,4\}$.
Their reference axes and heading linearization come from a straight nominal trajectory at 5\,m/s.
For nominal velocity $v_i$ and perturbation $\delta v_i$, the heading component is
$(v_{i,x}\delta v_{i,y}-v_{i,y}\delta v_{i,x})/\max(\|v_i\|_2^2,0.25)$.

Collect these templates in $A\in\mathbb R^{120\times6}$.
Their coordinates in the existing residual decoder are obtained by a regularized weighted least-squares solve,
\begin{equation}
 B_0=(M^\top Q_bM+\lambda_bI)^{-1}M^\top Q_bA,
 \label{eq:vital_basis}
\end{equation}
where $Q_b$ uses a 5\,m/rad heading scale and
$\lambda_b=10^{-6}\operatorname{mean}(\operatorname{diag}(M^\top Q_bM))$.
Whitening the decoded directions with their physical Gram matrix and a $1\%$ diagonal ridge
gives the fixed basis $B\in\mathbb R^{48\times6}$ in Eq.~\ref{eq:vital_rl_policy}. A common
amplitude sets the initial executed XY RMS to 0.2\,m using training observations and decoder
geometry. The final directions lie in the PCA span, where the templates can be applied through the residual decoder.

\independent{\textbf{Radial correction limiter.}}
For a candidate $r=Bq$, let $d=\operatorname{reshape}(Mr)\in\mathbb R^{40\times3}$.
The same scale limits each candidate's position, heading, and code magnitude without rotating its direction.
\begin{equation}
 a(r)=\max\!\left\{1,\frac{\max_i\|d_{i,xy}\|_2}{1\,\mathrm m},
       \frac{\max_i|d_{i,\psi}|}{0.2\,\mathrm{rad}},
       \frac{\|r\|_2}{4\sqrt{48}}\right\},\qquad
 \Pi_{\mathcal C}(r)=\frac{r}{a(r)}.
 \label{eq:vital_limiter}
\end{equation}
Before heading wrapping, $M\Pi_{\mathcal C}(r)$ adds the correction to the fixed scaffold and base residual.
The affine mean $\mu$ is counted once in the base trajectory, so a zero correction preserves the base prediction.

\independent{\textbf{Actor and prior.}}
The actor $q_\phi$ reads the base action expert's final state, ego and navigation input,
predicted tokens, and decoded trajectory. These have widths 4,096, 8, 304, and 120. A SiLU
network of shape $4528\!\to\!256\!\to\!256\!\to\!6$ has 1,226,758 trainable parameters.
We clip inputs normalized with fixed training statistics to $[-10,10]$. A zero output layer
starts the actor at the base plan, and the physical prior penalizes the size of its mean correction.
For $d_{\rm raw}=\operatorname{reshape}(MBq)$, define
\begin{equation}
 \Omega(q)=\frac{1}{120(0.15\,\mathrm m)^2}
  \sum_{i=1}^{40}
  \left(d_{{\rm raw},i,x}^2+d_{{\rm raw},i,y}^2
  +(5\,\mathrm{m/rad})^2d_{{\rm raw},i,\psi}^2\right).
 \label{eq:vital_prior}
\end{equation}
The prior in Eq.~\ref{eq:vital_rl_objective} penalizes the unbounded mean with $\beta=0.01$ on hard
scenes and $\eta=0.1$ on the anchors, giving the stronger penalty to cases where the base policy already scores well.

\independent{\textbf{Training reward and anchors.}}
The original training split supplies two disjoint pools. Reward optimization uses the 17,013
scenes with base PDMS below 90. The no-op prior uses 6,598 high-scoring anchors from a deterministic
subset selected before scoring. Both pools exclude validation and \texttt{navtest} logs.
The official non-reactive NAVSIM v1 scorer~\citep{dauner2024navsim} supplies $R_{\rm PDM}\in[0,1]$.
For $j$ in no-at-fault collision, drivable-area compliance, time-to-collision and driving-direction compliance, define
$c_j(s,\tau)=\max\{0,m_j(s,\tau_{\rm IL})-m_j(s,\tau)\}$.
Then
\begin{equation}
 R_\lambda(s,\tau)=R_{\rm PDM}(s,\tau)-\sum_j\lambda_jc_j(s,\tau).
 \label{eq:vital_native_reward}
\end{equation}
The four multipliers start at $0.5$.
After each actor update, an exponential average of deterministic-mean costs is updated with decay $0.95$, and
$\lambda\leftarrow\operatorname{clip}_{[0,2]}[\lambda+0.02(\bar c-0.001)]$.
Anchors contribute only the no-op prior, with no sampled reward rollouts.

\independent{\textbf{Antithetic Gaussian policy gradient.}}
For one training context, write
$F_\lambda(q)=R_\lambda(s,\tau_{\rm IL}+M\Pi_{\mathcal C}(Bq))$.
The Gaussian-smoothed reward is
$\bar F_\lambda(q)=\mathbb E_\xi F_\lambda(q+\sigma\xi)$.
With fixed multipliers and integrable rewards, the Gaussian score identity gives
\begin{equation}
 \nabla_q\bar F_\lambda(q)
 =\frac{1}{\sigma}\mathbb E_\xi[F_\lambda(q+\sigma\xi)\xi]
 =\frac{1}{2\sigma}\mathbb E_\xi
       [(F_\lambda(q+\sigma\xi)-F_\lambda(q-\sigma\xi))\xi].
 \label{eq:vital_score_identity}
\end{equation}
Gaussian symmetry gives the second equality. We estimate it with $K$ independent antithetic pairs.
\begin{equation}
 \hat g(q)=\frac{1}{2K\sigma}\sum_{k=1}^{K}
     [F_\lambda(q+\sigma\xi_k)-F_\lambda(q-\sigma\xi_k)]\xi_k .
 \label{eq:vital_antithetic}
\end{equation}
The estimator is unbiased for the gradient of the smoothed reward. Smoothing acts on the six
correction coordinates and uses score evaluations rather than differentiating through the scorer.
The training surrogate $-\langle q_\phi,\operatorname{sg}[\hat g(q_\phi)]\rangle$ passes the
estimated gradient to the actor, and the two quadratic priors are differentiated directly.

\independent{\textbf{Optimization and deployment.}}
Each update draws 16 hard scenes in shuffled uniform passes, uses $K=8$ pairs at $\sigma=1$,
and adds 64 no-op anchors. AdamW uses learning rate $10^{-4}$, weight decay $10^{-3}$, and
gradient clipping at 1. The 1,600 updates use 409,600 perturbed reward queries in addition to
deterministic-mean scoring. The frozen base generates ten-step plans during training. At every
inference budget we apply the same deterministic correction once using fixed normalization,
basis, and scale. The scorer is used only to optimize the actor during post-training.

\subsection{Derivations and Interpretation}
\label{app:method_analysis}

\independent{\textbf{Heading and units.}}
The affine decoder acts on positions and locally lifted headings before wrapping headings to
$[-\pi,\pi]$. For a pose vector $a$ in this common local chart, the trajectory norm in
Section~\ref{sec:vital_analysis} is $\|a\|_Q=(a^\top Qa)^{1/2}$ with $Q\succ0$. The heading
scale is 5\,m/rad in the actor prior. The lifted heading error bounds the shortest-arc difference
between the wrapped headings, allowing the same bound to hold after the decoder's final wrap.

\independent{\textbf{Reconstruction and prediction bounds.}}
With an orthonormal PCA basis and invertible scales, $Mz_r^\star=UU^\top(e^\star-\mu)$.
Subtracting the reconstruction gives Eq.~\ref{eq:vital_pca_remainder}. For the prediction
bound in Eq.~\ref{eq:vital_error}, assume the visual decoder is locally $L_v$-Lipschitz into
the fixed trajectory norm.
Next, add and subtract $D(z_v^\star,z_r^\star)$ to obtain
\begin{align}
 \tau-D(\hat z_v,\hat z_r)
 &=[\tau-D(z_v^\star,z_r^\star)]
   +[D_v(z_v^\star)-D_v(\hat z_v)]
   +M(z_r^\star-\hat z_r).
\end{align}
The triangle inequality combines the representation error with the two prediction terms. The visual term
uses $L_v$, and the residual term uses the induced norm $\|M\|_{2\to Q}=\sqrt{\lambda_{\max}(M^\top QM)}$.

\independent{\textbf{Conditional prediction depends on the representation.}}
Fix the conditioning map $h$.
Let $u=z^\star-\epsilon$, $X=(z_t,t,h(c))$, and $m(X)=\mathbb E[u\mid X]$.
Assume finite second moments and a fixed positive definite token weight $W$.
For the flow-matching term alone,
\begin{align}
 \mathbb E[w(t)\|v_\theta(X)-u\|_W^2]
 &=
 \mathbb E[w(t)\|v_\theta(X)-m(X)\|_W^2]\nonumber\\
 &\quad+
 \mathbb E\!\left[w(t)\operatorname{tr}
       \!\left(W\operatorname{Cov}[u\mid X]\right)\right].
 \label{eq:vital_conditional_risk}
\end{align}
\emph{Proof.}
Expand $v_\theta-u=(v_\theta-m)+(m-u)$ and condition on $X$.
The cross term vanishes because $\mathbb E[u-m\mid X]=0$.
The time weight is measurable with respect to $X$.
Where $w(t)>0$, the unconstrained population minimizer is therefore $m(X)$.
The conditional variance term describes the uncertainty left by the chosen target and
conditioning in the action objective. The probes in Appendix~\ref{app:representation_controls}
compare this mapping empirically within a fixed linear model class.

Decoding assigns physical meaning to a numerical token error.
Let $\widetilde D$ denote the composition of inverse token normalization, splitting, and
the trajectory decoder, as in Section~\ref{sec:vital_imitation}.
Locally, for differentiable $\widetilde D$ and small $\delta$,
\begin{equation}
 \|\widetilde D(z+\delta)-\widetilde D(z)\|_Q^2
 =\delta^\top J_{\widetilde D}(z)^\top QJ_{\widetilde D}(z)\delta
   +o(\|\delta\|_2^2).
 \label{eq:vital_pullback}
\end{equation}
In raw residual coordinates, the exact physical metric is $M^\top QM$.
For normalized tokens, the Jacobian additionally includes the inverse normalization map.
Two errors of the same standardized magnitude can therefore have different effects in physical
space, which is why token prediction error and decoded displacement can rank representations differently.

\independent{\textbf{Decoded one-step integration error.}}
Fix $h$ and an initial noise sample $z_0$.
Let $z_t$ solve $\dot z_t=v_\theta(z_t,t,h)$ and define the Euler endpoint
$z_{\mathrm E}=z_0+v_\theta(z_0,0,h)$.
Assume the learned field is continuously differentiable along a solution on $[0,1]$, and $\widetilde D$ is $L_D$-Lipschitz between its exact endpoint and the one-step output.
Define the material acceleration
$a(t)=\partial_t v_\theta(z_t,t,h)+J_zv_\theta(z_t,t,h)v_\theta(z_t,t,h)$.
Then
\begin{equation}
\begin{aligned}
 z_1-z_{\mathrm E}
 &=\int_0^1(1-t)a(t)\,dt,\\
 \|\widetilde D(z_1)-\widetilde D(z_{\mathrm E})\|_Q
 &\leq L_D\!\int_0^1(1-t)\|a(t)\|_2\,dt\\
 &\leq \frac{L_D}{2}\sup_{t\in[0,1]}\|a(t)\|_2 .
\end{aligned}
 \label{eq:vital_material_acceleration}
\end{equation}
\emph{Proof.}
The chain rule gives $d\,v_\theta(z_t,t,h)/dt=a(t)$.
Thus,
$z_1-z_{\mathrm E}=\int_0^1[v_\theta(z_t,t,h)-v_\theta(z_0,0,h)]\,dt
=\int_0^1\int_0^t a(s)\,ds\,dt$.
Interchanging the integrals yields the identity.
The triangle inequality and decoder Lipschitz continuity give the first bound in Eq.~\ref{eq:vital_material_acceleration}.
Taking the supremum of acceleration and using $\int_0^1(1-t)\,dt=1/2$ yields the final bound.

The one-step prediction error is bounded by $\|\tau-\widetilde D(z_1)\|_Q$ plus the first bound in
Eq.~\ref{eq:vital_material_acceleration}. Appendix~\ref{app:sampling_refinement} measures
prediction and integration separately using the recorded trajectories and an 80-step numerical reference.

\section{Experimental Protocol and Additional Results}
\label{app:experimental_protocol}

\subsection{Data, Evaluation, and Comparison Protocol}

\independent{\textbf{Data partitions.}}
Action learning uses 85,109 training scenes and a fixed, log-balanced validation panel of
4,096 scenes from 212 logs. Reconstruction uses all 18,179 validation scenes. Frozen-feature
probes use a nested panel of 512 scenes, while latency and solver diagnostics use 64 scenes.
The tokenizer, normalization statistics, residual basis, trajectory-only autoencoder, and residual
actor are fitted on the training split. Panel selection is independent of errors and reward
scores, and all methods and sampling budgets in a given comparison use the same evaluation scenes.

Trajectory error in the action-learning experiment is evaluated at the eight nominal planning times
$0.5,1.0,\ldots,4.0$ seconds. Clean-code reconstruction, ridge probes, and solver diagnostics instead
use all 40 points at 10\,Hz. Both grids span four seconds, but each weights the intervening trajectory samples differently.

\independent{\textbf{Metric components.}}
NAVSIM v1 reports PDMS and five components comprising no at-fault collision (NC), drivable-area
compliance (DAC), ego progress (EP), time-to-collision (TTC), and comfort (C).
NAVSIM v2 reports EPDMS with nine components comprising NC, DAC, driving-direction compliance
(DDC), traffic-light compliance (TLC), EP, TTC, lane keeping (LK), history comfort (HC), and
extended comfort (EC). All reported scores are scaled from 0 to 100.

\independent{\textbf{Planner configuration.}}
The planner uses $384\times672$ front images and the video/action backbone of
SimWAM~\citep{zhao2026simwam}. \method{} replaces its action target with four $[64\mid12]$
tokens from a separately trained, frozen tokenizer. Each token pairs visual motion with a metric
residual. The optional 1.23M-parameter actor applies one deterministic correction after decoding.
The NFE sweep varies the number of action flow evaluations while holding all weights and the inference seed fixed.

\independent{\textbf{Released checkpoint evaluation.}}
\label{app:released_baselines}
We evaluate the released SUV~\citep{yuan2026suv} and SimWAM~\citep{zhao2026simwam} imitation checkpoints on NAVSIM v1
\texttt{navtest}, using their native preprocessing and conditioning. SUV consumes four
front-camera observations, scene-dependent text, and 11 ego-state scalars. Its inference jointly
generates future visual tokens and actions. SimWAM uses the current front image, a fixed text
prompt, and eight ego/navigation scalars, with action-only inference. Their native image resolutions
are $384\times640$ and $384\times672$, respectively. Each checkpoint is evaluated at one and ten
sampling steps without retraining or candidate selection. Scores are computed independently for
each predicted trajectory against the PDM reference, using the same v1 metric cache and scorer
configuration as \method{}. Both use bfloat16 inference with native seeds 42 for SUV and
0 for SimWAM. Tab.~\ref{tab:vital_benchmark_navtest} uses a dagger for reproduced results and
includes \method{}'s imitation checkpoint for comparison.

\independent{\textbf{Published comparisons and score conventions.}}
Other results use the published settings in the cited papers. Camera counts refer to physical
views, including cameras stitched into a panorama. Frame counts cover current and past camera
observations used at inference, with n.r. marking counts not disclosed in the published configuration.

MeanFuser combines multiple proposals~\citep{wang2026meanfuser}, and DRIFT aggregates 48 proposals
in one pass~\citep{xing2026drift}. \method{} decodes a single plan directly, distinguishing
one action evaluation from one pass that aggregates a set of candidate plans.

We use the corrected one-stage v2 EPDMS reported by SUV~\citep{yuan2026suv} and the benchmark
protocol for two-stage \texttt{navhard}~\citep{cao2025pseudo}. Tables show scores to one decimal
place, displacement errors to two decimal places, and latency in whole milliseconds. Nonzero
displacements below $0.01$ use scientific notation. Gains and intervals use unrounded measurements.

\subsection{NAVSIM v2 Transfer Details}
\label{app:v2_transfer}
Formal evaluation follows the NAVSIM v2.2 \texttt{navhard} mapping without changing the v1-trained
model~\citep{cao2025pseudo}. The official mapping pairs original scenes with synthetic continuations,
and both stages use reactive traffic. Each mapping $m$ contains two adjacent original frames
indexed by $i$ and their synthetic continuations indexed by $j$. Let $s_{mi}^{(1)}$ and
$s_{mij}^{(2)}$ be their EPDMS values on $[0,1]$, and let $w_{mij}$ be the normalized
official proximity weights over the continuations, with $\sum_j w_{mij}=1$. For $M$ mappings,
the combined score reported on the 0--100 scale is
\begin{equation}
\operatorname{EPDMS}
= \frac{100}{2M}\sum_{m=1}^{M}\sum_{i=1}^{2}
s_{mi}^{(1)}\left(\sum_j w_{mij}s_{mij}^{(2)}\right).
\label{eq:vital_navhard_aggregation}
\end{equation}
The metric multiplies the two stage scores within each mapping before averaging. The components
use the same aggregation in Tab.~\ref{tab:vital_navhard_combined}, and Tab.~\ref{tab:vital_navhard_stages}
shows the separate stages alongside the combined scores. The imitation policy has closely matched
component scores at one and ten steps, consistent with the small difference in combined EPDMS.

\independent{\textbf{Inference budgets.}}
NFE counts iterative trajectory-generator evaluations, and plans counts candidate trajectories
before final selection. A dash denotes non-iterative prediction. GeoWAM forecasts future geometry
from three frames of eight camera views and uses a deterministic trajectory decoder.
PerceptDrive uses four frames from one front camera, predicts future features once, and integrates
its flow actor for 25 Euler steps to return one trajectory~\citep{lu2026geowam,liu2026perceptdrive}.
These models use future geometry or features to condition trajectory generation. \method{}
places visual motion in the action target and decodes it directly from current observations.
The one-stage v2 comparisons retain DriveSuprim with 256 candidates and SparseDriveV2 with
200 candidates~\citep{yao2026drivesuprim,sun2026sparsedrivev2}. DiffusionDriveV2 uses two
sampling steps over 20 proposals~\citep{zou2025diffusiondrivev2}. SGDrive and ReCogDrive use the
released five-step DDIM planners, each returning one trajectory. DriveFine's marked budget
comprises 12 masked-token sampling passes and one refinement pass~\citep{dang2026drivefine}.
GuideFlow's marked budget denotes 100 flow-integration steps over 256 anchors, with additional
constraint guidance and energy-based refinement~\citep{liu2026guideflow}. These counts extend beyond denoising.
In Tab.~\ref{tab:vital_benchmark_navhard}, n.r. marks unreported budgets and a dash identifies planners that predict trajectories directly.

As shown in Tab.~\ref{tab:vital_transfer_details}, one-step post-training improves Stage~1
EPDMS from 82.3 to 83.3, while Stage~2 moves from 47.1 to 46.2. The fixed scaffold gives the actor
room to adjust execution in the initial scene. Reactive continuations can shift the planning
problem from adjusting the path to choosing a new maneuver, the distinction illustrated in
Appendix~\ref{app:limitations}.

The one-stage diagnostic rescores the full \texttt{navtest} split in the v2 simulator with
non-reactive traffic and adjacent-frame extended comfort where available. Both imitation budgets
round to 90.7 EPDMS. Tab.~\ref{tab:vital_navtest_v2_components} shows that this agreement extends
to the component scores, rather than reflecting large gains and losses that cancel in the aggregate.

\begin{table}[!htbp]
\centering
\caption{\textbf{Residual post-training under NAVSIM v2 transfer.} For unchanged v1 checkpoints,
we report stage and combined EPDMS on reactive \texttt{navhard}, and corrected \texttt{navtest} EPDMS
with non-reactive traffic and extended comfort evaluated across adjacent frames.}
\label{tab:vital_transfer_details}
\begingroup\vitaltablefont\setlength{\tabcolsep}{3pt}\renewcommand{\arraystretch}{1.05}
\begin{tabularx}{.80\linewidth}{@{}L{1.6} C{.5} C{.95} C{.95} C{1.0} C{1.0}@{}}
\toprule
& & \multicolumn{3}{c}{Reactive \texttt{navhard}} & \texttt{navtest} \\
\cmidrule(lr){3-5}\cmidrule(l){6-6}
Model & NFE & Stage 1\vitalup & Stage 2\vitalup & Combined\vitalup & EPDMS\vitalup \\
\midrule
w/o residual RL & 1 & \vitalsecond{82.3} & \vitalbest{47.1} & \vitalsecond{38.5} & \vitalsecond{90.7} \\
\vitalrow \method{} & 1 & \vitalbest{83.3} & \vitalsecond{46.2} & \vitalbest{38.6} & \vitalbest{91.1} \\
w/o residual RL & 10 & \vitalsecond{82.3} & \vitalbest{47.3} & \vitalbest{38.6} & \vitalsecond{90.7} \\
\vitalrow \method{} & 10 & \vitalbest{82.9} & \vitalsecond{46.3} & \vitalsecond{38.1} & \vitalbest{91.2} \\
\bottomrule
\end{tabularx}
\endgroup
\end{table}

\begin{table}[!htbp]
\centering
\caption{\textbf{Corrected NAVSIM v2 \texttt{navtest} components.} The nine driving criteria accompany the aggregate EPDMS in Tab.~\ref{tab:vital_benchmark_transfer}. The \method{} rows use unchanged v1 checkpoints and the inference budgets and corrected v2 scoring protocol described in Appendix~\ref{app:v2_transfer}.}
\label{tab:vital_navtest_v2_components}
\begingroup
\fontsize{8}{9.2}\selectfont
\setlength{\tabcolsep}{1pt}
\renewcommand{\arraystretch}{1.04}
\begin{tabularx}{\linewidth}{@{}L{2.65} C{1.35} *{9}{C{.72}} C{1.52}@{}}
\toprule
Method & NFE/plans & NC & DAC & DDC & TLC & EP & TTC & LK & HC & EC & \textbf{EPDMS} \\
\midrule
\rowcolor{VitalLight}\multicolumn{12}{@{}l}{\textit{Regression / selection}}\\
DiffusionDriveV2 \vitaltablecitep{zou2025diffusiondrivev2} & 2/20 & 97.7 & 96.6 & 99.2 & \vitalsecond{99.8} & 88.9 & 97.2 & 96.0 & 97.8 & \vitalbest{91.0} & 87.5 \\
DriveSuprim~\vitaltablecitep{yao2026drivesuprim} & --/256 & 98.4 & 98.6 & \vitalsecond{99.6} & \vitalsecond{99.8} & \vitalsecond{90.5} & 97.8 & 97.0 & \vitalsecond{98.3} & 78.6 & 87.1 \\
SparseDriveV2~\vitaltablecitep{sun2026sparsedrivev2} & --/200 & 98.1 & 98.1 & \vitalsecond{99.6} & \vitalsecond{99.8} & \vitalbest{91.1} & 97.3 & 96.9 & 98.2 & 78.4 & 90.1 \\
\rowcolor{VitalLight}\multicolumn{12}{@{}l}{\textit{Vision--language}}\\
DriveVLA-W0~\vitaltablecitep{li2026drivevla} & AR/1 & 98.4 & 95.2 & 99.4 & \vitalbest{99.9} & 86.6 & 97.9 & 97.8 & \vitalsecond{98.3} & 82.7 & 86.9 \\
DriveWorld-VLA~\vitaltablecitep{liu2026driveworld} & n.r. & 98.6 & \vitalbest{99.1} & \vitalsecond{99.6} & \vitalsecond{99.8} & 87.4 & 97.9 & 97.0 & 97.8 & 78.6 & 86.8 \\
SGDrive~\vitaltablecitep{li2026sgdrive} & 5/1 & 98.6 & 94.3 & 99.5 & \vitalsecond{99.8} & 86.0 & 97.9 & 96.1 & \vitalsecond{98.3} & 85.9 & 86.2 \\
DriveFine~\vitaltablecitep{dang2026drivefine} & $13^{*}$/1 & \vitalsecond{98.7} & 97.3 & 99.5 & \vitalsecond{99.8} & 88.7 & 97.8 & 97.7 & \vitalbest{98.4} & 83.8 & 89.7 \\
\rowcolor{VitalLight}\multicolumn{12}{@{}l}{\textit{Predictive / world-action}}\\
EponaV2~\vitaltablecitep{xu2026eponav2} & n.r. & 98.5 & 97.4 & 99.5 & \vitalbest{99.9} & 87.9 & 98.1 & 97.7 & 98.2 & 77.4 & 88.9 \\
Metis~\vitaltablecitep{li2026metis} & 10/1 & 98.4 & 97.2 & \vitalsecond{99.6} & \vitalsecond{99.8} & 87.8 & 97.7 & 97.8 & \vitalbest{98.4} & 88.0 & 89.5 \\
Metis (Top-6)~\vitaltablecitep{li2026metis} & 10/6 & 98.5 & 97.5 & \vitalsecond{99.6} & \vitalsecond{99.8} & 87.9 & 97.8 & \vitalsecond{98.0} & \vitalbest{98.4} & \vitalsecond{90.0} & 90.3 \\
SUV@1~\vitaltablecitep{yuan2026suv} & 1/1 & -- & -- & -- & -- & -- & -- & -- & -- & -- & 89.8 \\
SUV@10 & 10/1 & \vitalbest{99.1} & 97.8 & \vitalbest{99.7} & \vitalsecond{99.8} & 87.8 & \vitalbest{98.7} & \vitalbest{98.1} & \vitalbest{98.4} & 88.3 & 91.0 \\
\rowcolor{VitalLight}\multicolumn{12}{@{}l}{\textit{\method{} with one generated trajectory}}\\
\method{} w/o RL@1 & 1/1 & 98.6 & 98.0 & \vitalsecond{99.6} & \vitalsecond{99.8} & 87.5 & 98.2 & 97.9 & \vitalbest{98.4} & 88.8 & 90.7 \\
\method{} w/o RL@10 & 10/1 & 98.6 & 98.1 & \vitalsecond{99.6} & \vitalsecond{99.8} & 87.5 & 98.2 & 97.9 & \vitalbest{98.4} & 88.7 & 90.7 \\
\vitalrow \textcolor{VitalPurple}{\textbf{\method{}@1}} & 1/1 & \vitalsecond{98.7} & 98.5 & \vitalbest{99.7} & \vitalsecond{99.8} & 87.8 & \vitalsecond{98.3} & 97.8 & \vitalbest{98.4} & 87.8 & \textcolor{VitalPurple}{\vitalsecond{91.1}} \\
\vitalrow \method{}@10 & 10/1 & 98.6 & \vitalsecond{98.7} & \vitalbest{99.7} & \vitalbest{99.9} & 87.9 & \vitalsecond{98.3} & 97.7 & \vitalbest{98.4} & 87.5 & \vitalbest{91.2} \\
\bottomrule
\end{tabularx}
\endgroup
\end{table}

\begin{table}[!htbp]
\centering
\caption{\textbf{Reactive NAVSIM v2 \texttt{navhard} results.} S1 and S2 show the nine driving components for each stage. The last column shows the official EPDMS after combining stages within each mapping, as in Tab.~\ref{tab:vital_benchmark_navhard}. Dashes in the component columns mark scores not reported by the source.}
\label{tab:vital_navhard_stages}
\begingroup
\fontsize{8}{9.2}\selectfont
\setlength{\tabcolsep}{1pt}
\renewcommand{\arraystretch}{1.02}
\begin{tabularx}{\linewidth}{@{}L{3.6} C{1.25} C{.75} *{9}{C{.7}} C{1.1}@{}}
\toprule
Method & NFE/plans & Stage & NC & DAC & DDC & TLC & EP & TTC & LK & HC & EC & \textbf{EPDMS} \\
\midrule
\rowcolor{VitalLight}\multicolumn{13}{@{}l}{\textit{Regression / selection}}\\
\multirow{2}{*}{LTF~\vitaltablecitep{chitta2022transfuser}} & \multirow{2}{*}{--/1} & S1 & 96.2 & 79.6 & 99.1 & 99.6 & 84.1 & 95.1 & 94.2 & 97.6 & 79.1 & \multirow{2}{*}{25.1} \\
 &  & S2 & 77.8 & 70.2 & 84.3 & 98.1 & 85.1 & 75.6 & 45.4 & 95.7 & \vitalbest{76.0} &  \\
\multirow{2}{*}{GuideFlow~\vitaltablecitep{liu2026guideflow}} & \multirow{2}{*}{$100^{*}$/256} & S1 & 96.6 & 80.5 & 96.3 & 99.3 & 82.3 & 94.9 & 91.5 & \vitalsecond{97.7} & 67.8 & \multirow{2}{*}{27.1} \\
 &  & S2 & \vitalbest{87.3} & 76.7 & \vitalbest{88.8} & \vitalbest{99.2} & 84.3 & \vitalbest{85.1} & 49.7 & 93.1 & 44.5 &  \\
\rowcolor{VitalLight}\multicolumn{13}{@{}l}{\textit{Vision--language}}\\
\multirow{2}{*}{DriveVLA-W0~\vitaltablecitep{li2026drivevla}} & \multirow{2}{*}{AR/1} & S1 & 96.8 & 83.3 & 99.0 & 99.6 & \vitalbest{84.6} & 95.3 & 96.4 & 97.6 & 78.2 & \multirow{2}{*}{24.4} \\
 &  & S2 & 76.8 & 64.3 & 79.9 & 98.3 & \vitalbest{89.2} & 75.0 & 46.8 & 95.8 & 53.1 &  \\
\multirow{2}{*}{ReCogDrive~\vitaltablecitep{xiong2026recogdrive}} & \multirow{2}{*}{5/1} & S1 & 96.4 & 78.9 & 98.7 & \vitalsecond{99.8} & 82.6 & 95.6 & 94.4 & 97.6 & 74.2 & \multirow{2}{*}{25.7} \\
 &  & S2 & 80.2 & 65.0 & 82.4 & 98.7 & 85.2 & 76.9 & 43.8 & \vitalbest{96.6} & 71.8 &  \\
\multirow{2}{*}{SGDrive~\vitaltablecitep{li2026sgdrive}} & \multirow{2}{*}{5/1} & S1 & 95.8 & 87.6 & 97.8 & \vitalsecond{99.8} & 84.4 & 94.7 & 92.9 & \vitalbest{97.8} & 28.9 & \multirow{2}{*}{25.5} \\
 &  & S2 & 79.4 & 65.4 & 79.1 & \vitalsecond{98.9} & \vitalsecond{88.9} & 75.3 & 42.7 & \vitalsecond{96.4} & 29.6 &  \\
\rowcolor{VitalLight}\multicolumn{13}{@{}l}{\textit{Predictive / world-action}}\\
\multirow{2}{*}{EponaV2~\vitaltablecitep{xu2026eponav2}} & \multirow{2}{*}{n.r.} & S1 & 97.3 & 90.7 & 99.4 & \vitalbest{100.0} & 83.3 & \vitalbest{97.3} & 97.3 & 97.6 & 60.9 & \multirow{2}{*}{36.1} \\
 &  & S2 & \vitalsecond{83.6} & 78.0 & \vitalsecond{88.0} & \vitalsecond{98.9} & 86.0 & \vitalsecond{80.3} & \vitalbest{50.1} & 96.1 & 52.0 &  \\
\multirow{2}{*}{Metis~\vitaltablecitep{li2026metis}} & \multirow{2}{*}{10/1} & S1 & 96.6 & 87.8 & 99.0 & 99.3 & \vitalsecond{84.5} & 95.6 & 97.8 & \vitalbest{97.8} & 77.8 & \multirow{2}{*}{32.2} \\
 &  & S2 & 79.6 & 73.3 & 84.9 & 97.8 & 85.8 & 76.6 & 47.7 & 95.4 & \vitalsecond{75.3} &  \\
\multirow{2}{*}{SUV@1~\vitaltablecitep{yuan2026suv}} & \multirow{2}{*}{1/1} & S1 & -- & -- & -- & -- & -- & -- & -- & -- & -- & \multirow{2}{*}{33.0} \\
 &  & S2 & -- & -- & -- & -- & -- & -- & -- & -- & -- & \\
\multirow{2}{*}{SUV@10} & \multirow{2}{*}{10/1} & S1 & 96.9 & \vitalsecond{94.2} & 99.3 & 99.6 & 84.1 & 95.6 & 96.7 & \vitalbest{97.8} & \vitalsecond{79.6} & \multirow{2}{*}{36.9} \\
 &  & S2 & 82.7 & 74.2 & 85.9 & 98.4 & 86.0 & 78.9 & 47.2 & 95.9 & 69.1 &  \\
\multirow{2}{*}{SimWAM@10~\vitaltablecitep{zhao2026simwam}} & \multirow{2}{*}{10/1} & S1 & \vitalbest{98.0} & 92.0 & \vitalbest{99.7} & 99.6 & 83.8 & 96.2 & 97.3 & \vitalbest{97.8} & 71.6 & \multirow{2}{*}{37.6} \\
 &  & S2 & 81.8 & 78.6 & 87.3 & 98.4 & 86.3 & 78.6 & 49.5 & 96.3 & 69.9 &  \\
\multirow{2}{*}{GeoWAM~\vitaltablecitep{lu2026geowam}} & \multirow{2}{*}{--/1} & S1 & \vitalsecond{97.7} & 91.5 & 99.1 & \vitalsecond{99.8} & 83.8 & 95.8 & 96.0 & \vitalbest{97.8} & 79.0 & \multirow{2}{*}{36.6} \\
 &  & S2 & 80.4 & 76.3 & 87.3 & 98.7 & \vitalsecond{88.9} & 76.2 & \vitalsecond{49.9} & 94.0 & 56.0 &  \\
\multirow{2}{*}{PerceptDrive~\vitaltablecitep{liu2026perceptdrive}} & \multirow{2}{*}{25/1} & S1 & 97.5 & \vitalsecond{94.2} & 99.0 & 99.4 & 83.9 & 95.6 & 96.4 & \vitalsecond{97.7} & 78.1 & \multirow{2}{*}{34.5} \\
 &  & S2 & 81.9 & 71.4 & 85.1 & 98.5 & 84.5 & 77.9 & 48.7 & \vitalbest{96.6} & 69.3 &  \\
\rowcolor{VitalLight}\multicolumn{13}{@{}l}{\textit{\method{} with one generated trajectory}}\\
\multirow{2}{*}{\method{} w/o RL@1} & \multirow{2}{*}{1/1} & S1 & 97.6 & 93.1 & 99.4 & 99.6 & 83.8 & 96.4 & \vitalsecond{98.7} & \vitalbest{97.8} & \vitalbest{80.0} & \multirow{2}{*}{\vitalsecond{38.5}} \\
 &  & S2 & 82.7 & 79.7 & 87.0 & 98.3 & 86.3 & 78.8 & 48.4 & 96.2 & 74.1 &  \\
\multirow{2}{*}{\method{} w/o RL@10} & \multirow{2}{*}{10/1} & S1 & 97.3 & 93.3 & 99.4 & 99.6 & 83.8 & 96.4 & \vitalbest{98.9} & \vitalbest{97.8} & \vitalbest{80.0} & \multirow{2}{*}{\vitalbest{38.6}} \\
 &  & S2 & 82.8 & \vitalbest{80.2} & 87.0 & 98.3 & 86.2 & 78.8 & 48.0 & 96.3 & 75.1 &  \\
\vitalrow &  & S1 & 97.6 & \vitalbest{94.4} & \vitalsecond{99.6} & 99.6 & 84.0 & 96.9 & \vitalbest{98.9} & \vitalbest{97.8} & 77.3 &  \\
\vitalrow \multirow{-2}{*}{\textcolor{VitalPurple}{\textbf{\method{}@1}}} & \multirow{-2}{*}{1/1} & S2 & 82.2 & 79.5 & 87.4 & 98.3 & 86.8 & 78.3 & 48.8 & 95.8 & 71.6 & \multirow{-2}{*}{\textcolor{VitalPurple}{\vitalbest{38.6}}} \\
\multirow{2}{*}{\method{}@10} & \multirow{2}{*}{10/1} & S1 & 97.0 & \vitalbest{94.4} & \vitalsecond{99.6} & \vitalsecond{99.8} & 84.2 & \vitalsecond{97.1} & \vitalsecond{98.7} & \vitalbest{97.8} & 76.4 & \multirow{2}{*}{38.1} \\
 &  & S2 & 82.3 & \vitalsecond{79.8} & 87.5 & 98.3 & 86.8 & 78.3 & 48.3 & 96.0 & 70.0 &  \\
\bottomrule
\end{tabularx}
\endgroup
\end{table}

\begin{table}[!htbp]
\centering
\caption{\textbf{Combined \texttt{navhard} components before residual RL.} The nine criteria use the EPDMS aggregation rule, which combines stages within each mapping before averaging. Component scores are closely matched at one and ten evaluations of the same v1 imitation checkpoint.}
\label{tab:vital_navhard_combined}
\begingroup
\fontsize{8}{9.2}\selectfont
\setlength{\tabcolsep}{1pt}
\renewcommand{\arraystretch}{1.1}
\begin{tabularx}{\linewidth}{@{}L{2.2} C{.8} *{9}{C{.8}} C{1.8}@{}}
\toprule
Model & NFE & NC & DAC & DDC & TLC & EP & TTC & LK & HC & EC & \textbf{EPDMS} \\
\midrule
\vitalrow \method{} w/o RL & 1 & \vitalbest{80.7} & \vitalsecond{74.4} & \vitalsecond{86.5} & \vitalbest{98.0} & \vitalbest{73.2} & \vitalsecond{76.4} & \vitalbest{48.0} & \vitalsecond{94.0} & \vitalsecond{60.0} & \vitalsecond{38.5} \\
\vitalrow \method{} w/o RL & 10 & \vitalsecond{80.6} & \vitalbest{74.9} & \vitalbest{86.7} & \vitalbest{98.0} & \vitalsecond{73.1} & \vitalbest{76.6} & \vitalsecond{47.6} & \vitalbest{94.1} & \vitalbest{60.4} & \vitalbest{38.6} \\
\bottomrule
\end{tabularx}
\endgroup
\end{table}

\subsection{Representation Controls}
\label{app:representation_controls}

\subsubsection{Frozen-Feature Predictability Probes}

\independent{\textbf{Trajectory-code control.}}
Trajectory AE provides a learned action code constructed from trajectories alone. It groups
the 40-point, 10\,Hz trajectory into four segments of ten
$(x,y,\theta)$ poses. A shared encoder maps the 30 coordinates of each segment through
$30\!\to\!256\!\to\!256\!\to\!76$ layers, with a symmetric SiLU decoder. The 186,474-parameter
AE achieves 1.80\,cm held-out reconstruction ADE and matches \method{}'s token shape.
The common interface holds the number and size of action tokens fixed while changing how they are constructed.

\begin{table}[!htbp]
\centering
\caption{\textbf{Probe fitting budgets.} E and V give ADE in meters for ego and navigation alone or with vision. MSE uses standardized V predictions.}
\label{tab:vital_alignment}
\begingroup\vitaltablefont\setlength{\tabcolsep}{2pt}\renewcommand{\arraystretch}{1.04}
\begin{tabularx}{.72\linewidth}{@{}L{1.6} C{0.8} C{0.85} C{0.85} C{0.9}@{}}
\toprule
Target & Train & E\vitaldown & V\vitaldown & MSE\vitaldown \\
\midrule
Trajectory AE & 819 & \vitalbest{1.09} & \vitalsecond{0.94} & \vitalbest{0.18} \\
\vitalrow \method{} & 819 & \vitalsecond{1.69} & \vitalbest{0.85} & \vitalsecond{0.40} \\
Trajectory AE & 2,048 & \vitalbest{1.09} & \vitalsecond{0.95} & \vitalbest{0.15} \\
\vitalrow \method{} & 2,048 & \vitalsecond{1.65} & \vitalbest{0.84} & \vitalsecond{0.47} \\
Trajectory AE & 8,192 & \vitalbest{1.08} & \vitalsecond{0.80} & \vitalbest{0.12} \\
\vitalrow \method{} & 8,192 & \vitalsecond{1.62} & \vitalbest{0.72} & \vitalsecond{0.38} \\
\bottomrule
\end{tabularx}
\endgroup
\end{table}

\independent{\textbf{Inputs and fitting.}}
Each probe uses 6,144 channelwise statistics from the frozen current-frame video prefix along with eight ego and navigation scalars.
Nested training sets of 819, 2,048, and 8,192 scenes share one fixed held-out panel. Ridge
regression predicts 304 standardized coordinates with penalty $0.01$, an unpenalized intercept,
and 1,870,512 fitted parameters per target. Within each budget, the design matrix is shared by both targets.

\independent{\textbf{Planning performance.}}
At the largest fitting budget, matched images raise \method{} by 23.0 PDMS points and
Trajectory AE by 16.0, as shown in Tab.~\ref{tab:vital_probe_pdms}. Tab.~\ref{tab:vital_alignment}
shows why the decoder matters for this comparison. \method{}'s higher standardized token MSE
corresponds to lower decoded displacement, so the two spaces rank the representations differently.

\independent{\textbf{Scene correspondence.}}
To test whether the gain comes from scene correspondence, we independently permute visual features
within the training and validation partitions while preserving ego states, targets, and input size.
As shown in Tab.~\ref{tab:vital_probe_shuffle}, shuffling lowers \method{} from 79.4 to 52.3 PDMS
at the same feature dimension. The loss of geometric accuracy indicates that the gain comes from
the correspondence between each image and its action target.
\begin{table}[!htbp]
\centering
\caption{\textbf{Probe planning scores.} V adds matched visual features to the ego and navigation input E. Gains use unrounded PDMS.}
\label{tab:vital_probe_pdms}
\begingroup\vitaltablefont
\setlength{\tabcolsep}{1.5pt}\renewcommand{\arraystretch}{1.04}
\begin{tabularx}{.72\linewidth}{@{}L{1.5} C{0.8} C{0.8} C{0.9}@{}}
\toprule
Target & E\vitalup & V\vitalup & Gain\vitalup \\
\midrule
Trajectory AE & \vitalbest{57.3} & \vitalsecond{73.2} & \vitalsecond{16.0} \\
\vitalrow \method{} & \vitalsecond{56.4} & \vitalbest{79.4} & \vitalbest{23.0} \\
\bottomrule
\end{tabularx}
\endgroup
\end{table}

\begin{table}[!htbp]
\centering
\caption{\textbf{Visual correspondence control.} Both probes use the same held-out scenes with matched or shuffled images. ADE is in meters.}
\label{tab:vital_probe_shuffle}
\begingroup\vitaltablefont
\setlength{\tabcolsep}{1.5pt}\renewcommand{\arraystretch}{1.04}
\begin{tabularx}{.72\linewidth}{@{}L{1.35} C{1.0} C{0.85} C{0.8}@{}}
\toprule
Target & Vision & ADE\vitaldown & PDMS\vitalup \\
\midrule
Trajectory AE & Matched & \vitalsecond{0.80} & \vitalsecond{73.2} \\
Trajectory AE & Shuffled & \vitalbest{1.38} & \vitalsecond{50.1} \\
\vitalrow \method{} & Matched & \vitalbest{0.72} & \vitalbest{79.4} \\
\vitalrow \method{} & Shuffled & \vitalsecond{2.20} & \vitalbest{52.3} \\
\bottomrule
\end{tabularx}
\endgroup
\end{table}

\begin{figure}[!htbp]
  \centering
  \includegraphics[width=.86\linewidth]{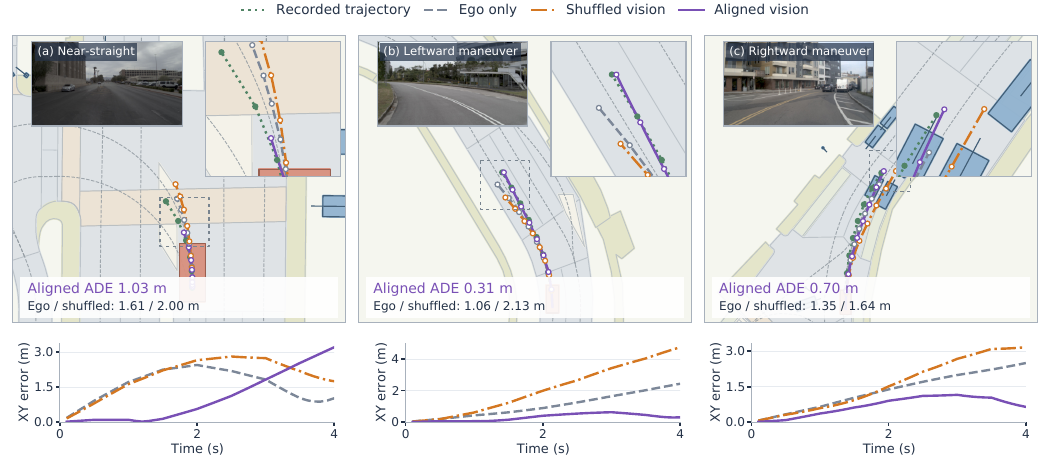}
  \caption{\textbf{Visual correspondence in probe predictions.} Lower panels show XY error for matched vision, ego input alone, and shuffled vision.}
  \label{fig:vital_conditioning_cases}
\end{figure}

\independent{\textbf{Visual correspondence in decoded predictions.}}
\label{app:qualitative_alignment}
Fig.~\ref{fig:vital_conditioning_cases} shows three \method{} probes fitted on 8,192 scenes.
Across near-straight motion and both turn directions, matching the image to the scene brings
the decoded path closer to the recorded maneuver than either ego input alone or shuffled visual features from another scene.

\subsubsection{Learning with a Randomly Initialized Action Expert}
\label{app:action_learning}

\independent{\textbf{Initialization and training exposure.}}
We compare raw trajectories, Trajectory AE, and \method{} under a shared initialization and
update budget to isolate the effect of the action target. Limiting training to one pass tests
how effectively each target learns the mapping from visual conditions to actions with the
same data exposure. Raw prediction retains SimWAM's eight $(x,y,\theta)$ poses at $0.5$-second intervals.
AE and \method{} predict $4\times76$ tokens and use their frozen codecs for decoding.
All three load the same pretrained video and proprioceptive foundation. Only the action expert
is randomly initialized. Its shared core tensors have identical initial values,
and target-specific projections map each trajectory or token interface into this common core.
All methods optimize the action expert and the same current-frame adapters on identical batches,
with no residual post-training.
Tab.~\ref{tab:learning_protocol} summarizes the shared optimization budget and inference settings.

\begin{table}[!htbp]
\centering
\caption{\textbf{Action-target ablation protocol.} All targets share the action-core initialization
and batch order for one training pass, including padding.}
\label{tab:learning_protocol}
\begingroup\vitaltablefont\setlength{\tabcolsep}{4pt}\renewcommand{\arraystretch}{1.1}
\begin{tabularx}{\linewidth}{@{}L{.9} C{1.1} L{.9} C{1.1}@{}}
\toprule
Setting & Value & Setting & Value \\
\midrule
Initialization seed & 42 & Global batch size & 20 \\
Updates / presentations & 4,256 / 85,120 & GPUs per method & Two 48\,GB RTX 4090s \\
Optimizer / peak LR & AdamW / $2\times10^{-5}$ & Warmup / schedule & 1,000 / cosine \\
Trainable modules & Action expert and current-frame adapters & Adapter scope & Last 8 blocks, rank 16 \\
Evaluation panel & Fixed and log-balanced & Inference seed / NFE & 0 / 1 and 10 \\
\bottomrule
\end{tabularx}
\endgroup
\end{table}

\independent{\textbf{Action objectives.}}
All targets use normalized action flow matching and video flow matching weighted by 0.25,
without auxiliary losses. The raw baseline predicts trajectory coordinates with unit action
weight, while AE predicts a trajectory-only code. For \method{}, visual and residual flow terms
have relative weights of one and two, normalized by their weighted coordinate count.
AE and \method{} recover poses through their respective frozen decoders. Tokenizers and
normalization statistics remain fixed throughout action learning, and the shared budget of
4,256 updates covers a full pass through the action-learning split after fitting the tokenizers and initializing the video foundation.

\independent{\textbf{Action-target comparison under matched updates.}}
As shown in Tab.~\ref{tab:vital_convergence}, AE's advantage over raw prediction emerges by
3,000 updates and reaches 9.9 one-step PDMS points after the shared pass. Learned trajectory
coding therefore provides a strong control for the effect of the representation. As shown in
Fig.~\ref{fig:vital_action_learning}b, \method{} gains a further 5.1 points with the same token
shape, initialization, and training exposure. Combined with the visual correspondence probes,
this gain supports the role of visual grounding in making action supervision easier to learn
from the available context, complementing the matched visual probe while controlling for token shape and training exposure.
\begin{table}[!htbp]
\centering
\caption{\textbf{Trajectory-only learning progress.} $\Delta$ is AE minus raw PDMS, with confidence intervals computed by paired resampling of validation logs.}
\label{tab:vital_convergence}
\begingroup\vitaltablefont
\setlength{\tabcolsep}{0.7pt}\renewcommand{\arraystretch}{1.04}
\begin{tabularx}{.72\linewidth}{@{}C{0.95327} C{1.00935} C{0.72897} C{0.78505} C{0.78505} C{1.73831}@{}}
\toprule
Updates & Seen & Raw\vitalup & AE\vitalup & $\Delta$\vitalup & 95\% CI \\
\midrule
1,000 & 20,000 & 45.6 & 47.6 & +2.0 & $[-0.1,4.0]$ \\
2,000 & 40,000 & 52.2 & 53.3 & +1.1 & $[-0.9,3.0]$ \\
3,000 & 60,000 & 55.2 & 64.5 & +9.3 & $[7.2,11.4]$ \\
4,000 & 80,000 & 55.0 & 65.0 & +10.0 & $[8.0,12.0]$ \\
4,256 & 85,120 & 55.0 & 64.9 & +9.9 & $[8.0,12.0]$ \\
\bottomrule
\end{tabularx}
\endgroup
\end{table}

\begin{table}[!htbp]
\centering
\caption{\textbf{Trajectory-only score contributions.} Progress, TTC, and comfort sum to PDMS after applying the NC and DAC gates.}
\label{tab:vital_score_terms}
\begingroup\vitaltablefont
\setlength{\tabcolsep}{0.8pt}\renewcommand{\arraystretch}{1.04}
\begin{tabularx}{.72\linewidth}{@{}L{1.79439} C{0.61682} C{0.8972}C{0.8972}C{0.8972}C{0.89719}@{}}
\toprule
Target & NFE & EP\vitalup & TTC\vitalup & C\vitalup & PDMS\vitalup \\
\midrule
Raw & 1 & \vitalsecond{24.4} & \vitalsecond{25.7} & \vitalsecond{4.9} & \vitalsecond{55.0} \\
Raw & 10 & \vitalbest{26.7} & \vitalbest{32.3} & \vitalsecond{11.6} & \vitalbest{70.6} \\
AE & 1 & \vitalbest{25.1} & \vitalbest{27.7} & \vitalbest{12.1} & \vitalbest{64.9} \\
AE & 10 & \vitalsecond{25.2} & \vitalsecond{28.7} & \vitalbest{12.5} & \vitalsecond{66.3} \\
\bottomrule
\end{tabularx}
\endgroup
\end{table}

\independent{\textbf{Planning gains across sampling budgets.}}
Tab.~\ref{tab:vital_action_sampling} evaluates the same final weights at one and ten steps.
\method{} scores 70.1 and 71.4 PDMS, exceeding AE's 64.9 and 66.3 at both budgets.
Raw prediction improves by 15.6 points with ten evaluations, while AE and \method{} each gain 1.4.
Thus both learned token representations support small sampling gaps, while \method{} achieves
the highest PDMS at both budgets. At ten steps AE has lower displacement errors, while
\method{} retains higher collision avoidance, road compliance, and progress scores. The
two metrics thus distinguish how closely a trajectory follows recorded motion from how well it performs under simulated driving criteria.
\begin{table}[!htbp]
\centering
\caption{\textbf{Final-checkpoint planning at matched solver budgets.} All targets train for one pass and retain the same weights at both budgets. ADE and FDE use eight points over four seconds, and the DDC column is reported separately from the v1 PDMS calculation.}
\label{tab:vital_action_sampling}
\begingroup\vitaltablefont\setlength{\tabcolsep}{2pt}\renewcommand{\arraystretch}{1.04}
\begin{tabularx}{\linewidth}{@{}L{1.65096} C{0.61421} C{0.94924} C{0.94924} C{0.94924} C{0.94924} C{0.94924} C{0.94924} C{1.06091} C{1.02924} C{0.94924}@{}}
\toprule
Target & NFE & NC\vitalup & DAC\vitalup & DDC\vitalup & EP\vitalup & TTC\vitalup & C\vitalup & PDMS\vitalup & ADE (m)\vitaldown & FDE (m)\vitaldown \\
\midrule
Raw trajectory & 1 & 92.5 & \vitalsecond{80.3} & 94.6 & 58.6 & 79.1 & 35.2 & 55.0 & 1.46 & 3.49 \\
Trajectory AE & 1 & \vitalsecond{95.4} & 76.8 & \vitalsecond{95.0} & \vitalsecond{60.3} & \vitalsecond{87.0} & \vitalbest{99.6} & \vitalsecond{64.9} & \vitalsecond{1.25} & \vitalsecond{2.45} \\
\vitalrow \method{} & 1 & \vitalbest{97.1} & \vitalbest{90.0} & \vitalbest{97.3} & \vitalbest{71.2} & \vitalbest{87.7} & \vitalsecond{51.6} & \vitalbest{70.1} & \vitalbest{1.16} & \vitalbest{2.13} \\
\midrule
Raw trajectory & 10 & 95.2 & \vitalsecond{87.2} & \vitalsecond{97.4} & \vitalsecond{64.2} & \vitalsecond{89.4} & \vitalsecond{84.0} & \vitalsecond{70.6} & \vitalbest{0.99} & 2.82 \\
Trajectory AE & 10 & \vitalsecond{95.9} & 78.8 & 96.2 & 60.7 & 87.7 & \vitalbest{99.2} & 66.3 & \vitalsecond{1.10} & \vitalbest{2.22} \\
\vitalrow \method{} & 10 & \vitalbest{98.2} & \vitalbest{89.4} & \vitalbest{97.6} & \vitalbest{69.8} & \vitalbest{91.0} & 57.4 & \vitalbest{71.4} & 1.20 & \vitalsecond{2.44} \\
\bottomrule
\end{tabularx}
\endgroup
\end{table}

\Needspace{3\baselineskip}
\independent{\textbf{Evidence from fully trained planners.}}
The released SimWAM imitation checkpoint~\citep{zhao2026simwam} provides a fully trained
trajectory-target reference in Tab.~\ref{tab:vital_benchmark_navtest}. Its 90.1 PDMS at ten
steps establishes strong planning quality, while its 66.6 at one step shows that extensive
training still leaves a large sampling gap. \method{} reaches 90.2 and 90.3 with the same
backbone at one and ten steps before residual RL. These fully trained planners complement
the common-budget ablation. The visual-only planner in the same table already reaches 90.0
PDMS at one step. Adding the jointly predicted metric residual brings it to 90.2 and raises
both road compliance and progress. Thus the visual target accounts for most of the one-step
performance, while the metric residual improves road compliance and progress with the actor turned off.

\begin{figure}[!htbp]
  \centering\includegraphics[width=.86\linewidth]{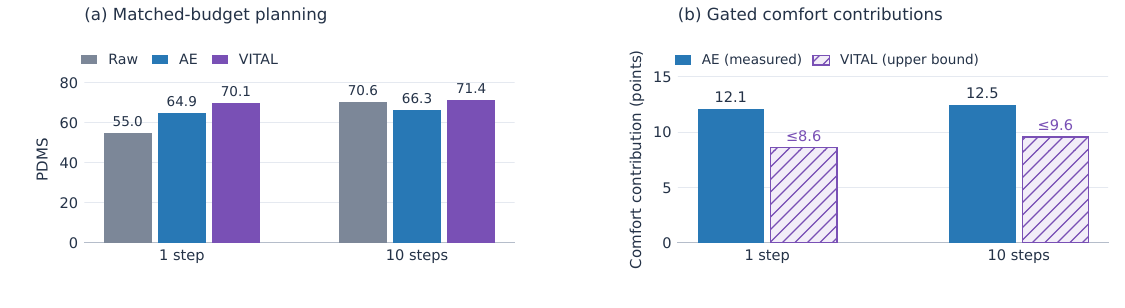}
  \caption{\textbf{Planning gains and comfort.} PDMS at one and ten evaluations appears on the left.
  On the right, solid bars give the measured comfort contribution for AE and hatched bars give
  the upper bound for \method{} under the same scoring rule in Eq.~\ref{eq:vital_comfort_bound}.}
  \label{fig:vital_action_diagnostics}
\end{figure}

\independent{\textbf{Planning score decomposition.}}
The component scores distinguish the tokenizer's planning gains from comfort improvements.
AE reaches 99.6/99.2 comfort at one/ten steps, compared with 51.6/57.4 for \method{}.
Yet \method{} improves NC, DAC, EP, TTC, and DDC at both budgets. This advantage also survives
the multiplicative gating in PDMS. For per-scene components in $[0,1]$, write
\begin{equation}
\begin{aligned}
 \mathrm{PDMS}&=100\,\mathbb E\!\left[
 \mathrm{NC}\,\mathrm{DAC}\,
 \frac{5\mathrm{EP}+5\mathrm{TTC}+2\mathrm C}{12}\right],\\
 G_C&=\frac{100}{6}\mathbb E[\mathrm{NC}\,\mathrm{DAC}\,\mathrm C]
 \leq\frac{\overline C}{6},
\end{aligned}
 \label{eq:vital_comfort_bound}
\end{equation}
where $\overline C=100\mathbb E[\mathrm C]$ is the reported comfort score.
As shown in Tab.~\ref{tab:vital_score_terms}, comfort contributes 7.3 points of AE's 9.9-point
one-step improvement over raw prediction. For \method{}, even the upper bounds on $G_C$,
8.6/9.6 points, are below AE's measured 12.1/12.5 points.
\method{}'s higher PDMS must therefore come from a larger combined gated progress and TTC contribution
despite a smaller comfort contribution. Fig.~\ref{fig:vital_action_diagnostics} shows this
contrast between the comfort contribution and the combined contribution of progress and TTC after gating for collision and road compliance.

\independent{\textbf{Evaluation protocol.}}
All evaluations use the same validation panel, with initial noise fixed across sampling budgets.
Confidence intervals use 20,000 paired bootstrap replicates over validation logs, retaining all
scenes within each sampled log and keeping the same scenes paired across the two models or budgets in each comparison.

\subsubsection{Decoder Sensitivity}
\begin{figure}[!htbp]
  \centering
  \includegraphics[width=.50\linewidth]{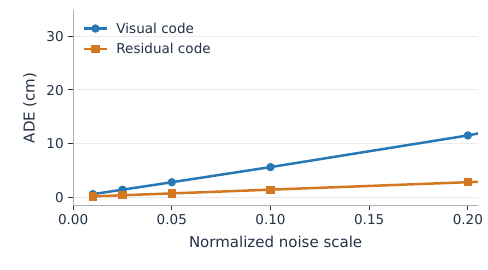}
  \caption{\textbf{Sensitivity to normalized token noise.} Each curve varies the noise on one group while fixing the other to its clean tokens.}
  \label{fig:vital_decoder_sensitivity}
\end{figure}

We perturb one normalized token group with Gaussian noise while fixing the other at its clean
value. In Fig.~\ref{fig:vital_decoder_sensitivity}, $\sigma_z=0.10$ gives 5.61\,cm ADE for visual
tokens and 1.39\,cm for residual tokens. Equal noise scales in the normalized token space
therefore produce different amounts of displacement after decoding to the same trajectory grid.

\subsection{Inference Efficiency and Solver Diagnostics}

\independent{\textbf{Timing protocol.}}
Batch-one latency is measured on an exclusively reserved RTX 4090 after eight warmup calls. Inputs
are front images and ego metadata resident in host memory. The measurement includes preprocessing,
host-to-device transfer, image encoding, visual conditioning, action integration, and trajectory
construction. Scene-file I/O and initial weight loading are excluded. Scene-dependent visual
features are recomputed for every call. We synchronize CUDA around the complete call and report the
median and 95th percentile across the same scenes for all sampling budgets.

As shown in Tab.~\ref{tab:vital_sampling_values}, the one-step median is 149\,ms with the
actor and 150\,ms without it. For the final policy, ten steps take 490\,ms versus 149\,ms
at one step. The imitation policy gains 0.1 PDMS over the same sweep. Reusing visual
features makes the number of action evaluations the main source of the change in total planning latency.

\independent{\textbf{Numerical refinement.}}
\label{app:sampling_refinement}
Using the held-out solver panel and identical initial noise, we compare NFE 1, 10, 40, and 80 in decoded
trajectory space. As shown in Tab.~\ref{tab:vital_solver}, disagreement with the 80-step numerical
reference falls from 6.61\,cm at one step to 0.97\,cm at forty steps. ADE against recorded motion
stays within 63.10 to 63.40\,cm. The finer solver therefore improves integration on this panel
without a corresponding improvement in trajectory prediction.
\begin{table}[!htbp]
\centering
\caption{\textbf{Sampling quality and latency.} PDMS uses v1 \texttt{navtest}. Median and P95 report batch-one inference latency in milliseconds.}
\label{tab:vital_sampling_values}
\begingroup\vitaltablefont
\setlength{\tabcolsep}{1pt}\renewcommand{\arraystretch}{1.04}
\begin{tabularx}{.72\linewidth}{@{}L{1.6} C{0.6} C{0.8} C{1.0} C{1.0}@{}}
\toprule
Model & NFE & PDMS\vitalup & Median\vitaldown & P95\vitaldown \\
\midrule
w/o residual RL & 1 & \vitalsecond{90.2} & \vitalsecond{150} & \vitalbest{155} \\
\vitalrow \method{} & 1 & \vitalbest{90.8} & \vitalbest{149} & \vitalsecond{157} \\
w/o residual RL & 2 & \vitalsecond{90.2} & \vitalbest{184} & \vitalsecond{193} \\
\vitalrow \method{} & 2 & \vitalbest{90.9} & \vitalsecond{187} & \vitalbest{192} \\
w/o residual RL & 4 & \vitalsecond{90.2} & \vitalsecond{262} & \vitalbest{270} \\
\vitalrow \method{} & 4 & \vitalbest{91.0} & \vitalbest{261} & \vitalbest{270} \\
w/o residual RL & 10 & \vitalsecond{90.3} & \vitalbest{487} & \vitalbest{495} \\
\vitalrow \method{} & 10 & \vitalbest{91.0} & \vitalsecond{490} & \vitalsecond{502} \\
\bottomrule
\end{tabularx}
\endgroup
\end{table}

\begin{table}[!htbp]
\centering
\caption{\textbf{Numerical solver refinement.} Errors are in centimeters. The last column gives ADE relative to the NFE 80 prediction.}
\label{tab:vital_solver}
\begingroup\vitaltablefont
\setlength{\tabcolsep}{1pt}\renewcommand{\arraystretch}{1.04}
\begin{tabularx}{.72\linewidth}{@{}C{0.61905} C{1.04762} C{1.04762} C{1.28571}@{}}
\toprule
NFE & GT ADE\vitaldown & GT FDE\vitaldown & To NFE 80\vitaldown \\
\midrule
1 & 63.30 & 166.00 & 6.61 \\
10 & 63.10 & 167.00 & 3.60 \\
40 & 63.40 & 167.00 & 0.97 \\
80 & 63.40 & 167.00 & 0.00 \\
\bottomrule
\end{tabularx}
\endgroup
\end{table}

\begin{figure}[!htbp]
  \centering
  \includegraphics[width=.86\linewidth]{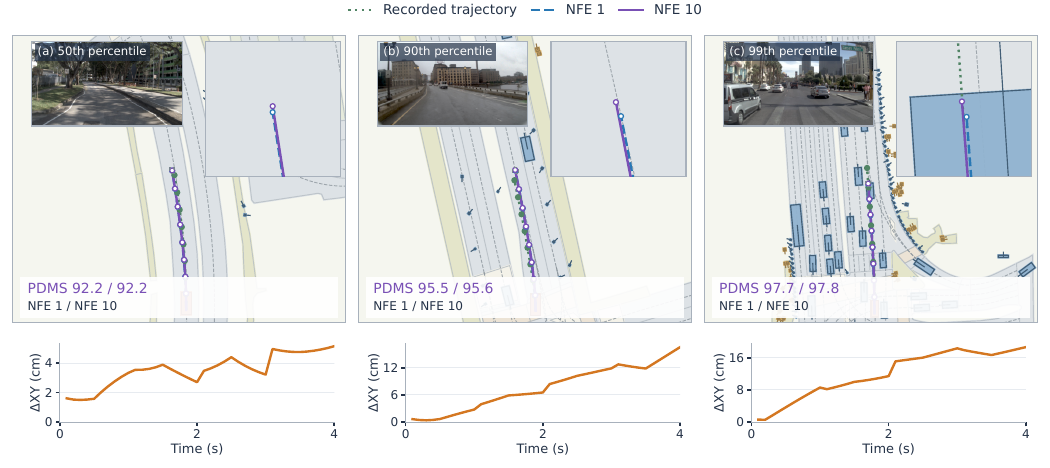}
  \caption{\textbf{One-step and ten-step trajectories.} Insets show the 50th, 90th, and 99th percentiles of XY change among high-scoring plans.}
  \label{fig:vital_sampling_cases}
\end{figure}

\independent{\textbf{One-step and ten-step examples.}}
\label{app:qualitative_sampling}
Fig.~\ref{fig:vital_sampling_cases} holds the policy and inference seed fixed across budgets.
Even where the insets show larger local differences, the paths retain the same maneuver.
In all three cases, additional evaluations change local path geometry while the resulting PDMS shifts by at most 0.1 points.

\subsection{Residual-Policy Controls}
\begin{figure}[!htbp]
  \centering
  \includegraphics[width=.50\linewidth]{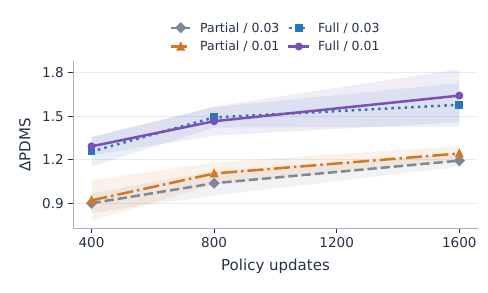}
  \caption{\textbf{PDMS gains across policy updates.} Bands show one standard deviation around the mean across three training seeds.}
  \label{fig:additional_controls}
\end{figure}

As shown in Tab.~\ref{tab:vital_rl} and Fig.~\ref{fig:additional_controls}, expanding the hard
scene pool from 13,105 to 17,013 improves the mean gain by about 0.4 PDMS points across three
seeds. At the tested prior weights, exposing the actor to more hard scenes has a larger effect
than changing the strength of the correction penalty applied to the same hard scene pool.
\begin{table}[!htbp]
\centering
\caption{\textbf{Reward-pool and prior controls.} PDMS gain and standard deviation (SD) use three seeds after 1,600 updates at NFE 10.
All four settings use the same no-op anchors, token normalization, correction geometry, and budget for evaluating candidate rewards.}
\label{tab:vital_rl}
\begingroup\vitaltablefont
\setlength{\tabcolsep}{1pt}\renewcommand{\arraystretch}{1.04}
\begin{tabularx}{.72\linewidth}{@{}C{1.05} C{1.15} C{0.7} C{1.25} C{0.85}@{}}
\toprule
Pool & Scenes & Prior & $\Delta$PDMS\vitalup & SD \\
\midrule
Partial & 13,105 & 0.03 & \vitalsecond{1.2} & 0.04 \\
Partial & 13,105 & 0.01 & \vitalsecond{1.2} & 0.05 \\
Full & 17,013 & 0.03 & \vitalbest{1.6} & 0.15 \\
\vitalrow Full & 17,013 & 0.01 & \vitalbest{1.6} & 0.18 \\
\bottomrule
\end{tabularx}
\endgroup
\end{table}

\begin{figure}[!htbp]
  \centering\includegraphics[width=.86\linewidth]{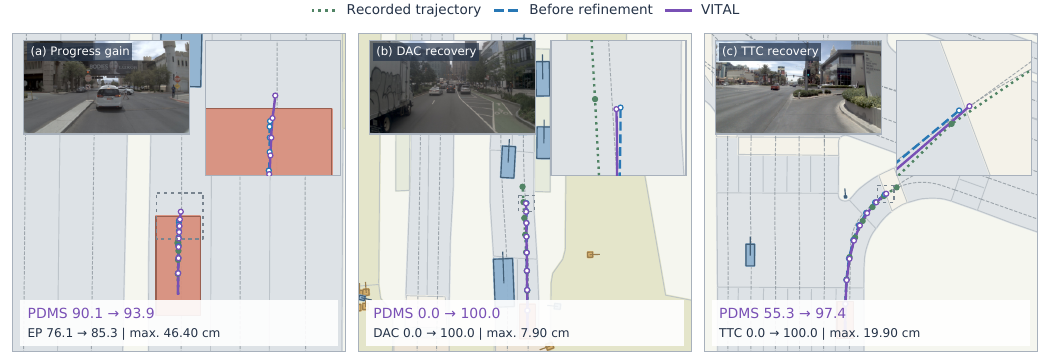}
  \caption{\textbf{One-step residual refinement.} Insets show corrections that improve progress in a, road compliance in b, and TTC in c.}
  \label{fig:vital_cases}
\end{figure}

\independent{\textbf{Local corrections in driving scenes.}}
\label{app:qualitative_residual}
Fig.~\ref{fig:vital_cases} illustrates how a fixed maneuver can admit different local corrections.
In panel b, a maximum displacement of 7.9\,cm restores road compliance and raises PDMS from
0.0 to 100.0. Near a road boundary, this links reconstruction precision in
Appendix~\ref{app:qualitative_codec} to a concrete change in the simulated execution score.

\subsection{Action Tokenizer Transfer Across Planners}
\label{app:cross_planner_transfer}

\independent{\textbf{Shared action representation.}}
We adapt the released imitation checkpoints of DiffusionDrive~\citep{liao2025diffusiondrive}
and DriveVA~\citep{liu2026driveva} on the NAVSIM training split without RL. Both replace eight
waypoints with the action representation from Appendix~\ref{app:method_details}. Each of the four
tokens pairs $d_v=64$ visual motion coordinates with $d_r/L=12$ metric residual coordinates.
Token normalization uses training labels, and one frozen codec decodes 40 poses over four seconds at 10\,Hz.

\independent{\textbf{DiffusionDrive.}}
We retain the camera and LiDAR backbone, bird's-eye view and object supervision, and 20 scored
proposals. Each proposal predicts action tokens rather than waypoints. We map each trajectory
anchor to token space using the label of the nearest expert trajectory in the training split.
Bird's-eye view attention still reads decoded positions sampled at 2\,Hz, so the model queries
scene features in physical coordinates. We retain anchor assignment and classification and
supervise decoded positions and headings. The scorer learns from soft labels based on candidate
trajectory errors. We retain the truncated DDIM schedule and remove waypoint clipping for the
standardized tokens. The backbone, auxiliary heads, and adapted planning heads jointly update
60.67M parameters, with the old waypoint branch and the codec frozen.

\independent{\textbf{DriveVA.}}
We adapt the action encoder and output head to predict the same four tokens while retaining the ego
speed prefix, joint video and action denoising in the Wan DiT, and the video flow matching objective.
Standardized tokens replace clipped waypoint coordinates. We train the action encoder, output
head, and rank-8 attention adapters in the final four blocks, totaling 12.86M trainable parameters.
The base DiT, VAE, text encoder, and shared action codec remain fixed during this adaptation.

\independent{\textbf{Evaluation protocol.}}
Tab.~\ref{tab:vital_cross_planner} compares each adapted model with its released checkpoint
on the same scenes at one and two denoising evaluations. We use four-second non-reactive
PDM simulation on NAVSIM v1.1 \texttt{navtest}~\citep{dauner2024navsim}. DiffusionDrive covers
all 12,146 scenes in FP32, and DriveVA covers all 12,123 scenes with the required five-frame
history in BF16. DriveVA uses seed zero and jointly denoises video and actions while skipping final RGB decoding.

\Needspace{4\baselineskip}
\independent{\textbf{Latency measurement.}}
We time batch-one inference on an RTX 4090 without PDM scoring. DiffusionDrive starts from
features already on the GPU. DriveVA starts from history images in host memory and includes
text encoding, history VAE encoding, and joint denoising, with image I/O and RGB decoding excluded.
We calibrate original checkpoints on the same scene with five warmup calls followed by
30 timed calls and CUDA synchronization around the measured inference.

\begin{table}[!htbp]
\centering
\caption{\textbf{Action tokenizer transfer on NAVSIM v1.1 \texttt{navtest}.} Shaded rows use
the same frozen \method{} codec. Both targets are evaluated on the same scenes within each
planner. Latency is reported in milliseconds using the model-specific timing scopes in
Appendix~\ref{app:cross_planner_transfer}.}
\label{tab:vital_cross_planner}
\begingroup\vitaltablefont
\setlength{\tabcolsep}{1.5pt}
\renewcommand{\arraystretch}{1.07}
\begin{tabularx}{\linewidth}{@{}L{1.62} C{.56} C{.96} *{5}{C{.81}} *{2}{C{.98}} *{2}{C{1.425}}@{}}
\toprule
& & \multicolumn{6}{c}{Planning scores} & \multicolumn{2}{c}{Displacement (m)}
& \multicolumn{2}{c}{Latency (ms)} \\
\cmidrule(lr){3-8}\cmidrule(lr){9-10}\cmidrule(l){11-12}
Action target & NFE & PDMS\vitalup & NC\vitalup & DAC\vitalup & EP\vitalup & TTC\vitalup & C\vitalup & ADE\vitaldown & FDE\vitaldown & Mean\vitaldown & P95\vitaldown \\
\midrule
\rowcolor{VitalLight}\multicolumn{12}{@{}l}{\textit{DiffusionDrive}} \\
Waypoints & 1 & 88.1 & 98.2 & 96.3 & 82.2 & 94.8 & 100.0 & 0.64 & 1.57 & 15 & 16 \\
\vitalrow \method{} tokens & 1 & \vitalbest{88.5} & 98.4 & 95.7 & 82.5 & 96.5 & 100.0 & 0.72 & 1.63 & 53 & 75 \\
Waypoints & 2 & 88.1 & 98.2 & 96.3 & 82.2 & 94.8 & 100.0 & 0.64 & 1.57 & 19 & 19 \\
\vitalrow \method{} tokens & 2 & \vitalbest{88.5} & 98.8 & 94.9 & 82.6 & 96.9 & 100.0 & 0.66 & 1.50 & 66 & 86 \\
\midrule
\rowcolor{VitalLight}\multicolumn{12}{@{}l}{\textit{DriveVA}} \\
Waypoints & 1 & 2.8 & 37.2 & 18.5 & 4.7 & 16.3 & 0.0 & 8.62 & 16.28 & 305 & 306 \\
\vitalrow \method{} tokens & 1 & \vitalbest{90.0} & 98.8 & 96.5 & 84.4 & 96.9 & 100.0 & 0.69 & 1.58 & 298 & 303 \\
Waypoints & 2 & \vitalbest{90.2} & 99.2 & 97.7 & 83.8 & 96.2 & 99.9 & 0.65 & 1.43 & 440 & 442 \\
\vitalrow \method{} tokens & 2 & 90.1 & 99.2 & 96.7 & 84.7 & 97.3 & 100.0 & 0.65 & 1.51 & 434 & 440 \\
\bottomrule
\end{tabularx}
\endgroup
\end{table}

\independent{\textbf{Token targets for scored proposals.}}
DiffusionDrive already reaches 88.1 PDMS at both budgets. Using the same 20 proposals with token
targets adds 0.4 PDMS at either budget, with higher TTC and lower DAC. One step cuts mean latency
from 66 to 53\,ms for the adapted planner, while the original two-step feature-to-plan time is 19\,ms.

\independent{\textbf{Token targets in joint video and action generation.}}
DriveVA reaches 90.0 PDMS in one step, compared with 2.8 for the original checkpoint at one
step and 90.2 at two. Relative to the original two-step setting, the adapted one-step model
reduces mean latency from 440 to 298\,ms and gives 0.69\,m ADE versus 0.65\,m.
The result preserves DriveVA's joint denoising pathway while bringing the action prediction close
to its original two-step planning quality. Together the two adaptations show that the same
codec can support both trajectory proposal scoring and joint video and action generation.

\FloatBarrier
\section{Comparison Protocol and Limitations}
\label{app:qualitative}

\subsection{Protocol for the Cross-Model Comparison}
\label{app:qualitative_comparison}

Fig.~\ref{fig:vital_wam_comparison} compares four-second plans from released DiffusionDrive~\citep{liao2025diffusiondrive},
SUV~\citep{yuan2026suv}, SimWAM-RL~\citep{zhao2026simwam}, and \method{} on the same NAVSIM v1
scene. DiffusionDrive uses three front-facing cameras and LiDAR and samples 20 proposals in two
DDIM steps. SUV uses four front images, scene-specific text, and 11 ego scalars to sample future
visual tokens jointly with actions. SimWAM-RL and \method{} use one front image and eight ego and
navigation scalars. All models use seed zero. With bfloat16 inference, \method{} uses one step and
SUV and SimWAM-RL use ten, while DiffusionDrive retains its native float32 implementation.

We score every prediction against the same PDM reference and compute ADE and FDE at eight times
from 0.5 to 4.0 seconds. Bird's-eye panels use the same viewport, insets, and recorded scene overlays.

In the curve shown in Fig.~\ref{fig:vital_wam_comparison}, \method{} follows the recorded path
more closely and reaches 100.0 PDMS. The shared insets show where SUV and SimWAM-RL overshoot
the recorded endpoint. SUV jointly samples video and actions, while SimWAM-RL predicts waypoints.
In contrast, one action evaluation of \method{} predicts the visual motion tokens and the paired metric residual used to decode the final path.

\subsection{Additional Cross-Model Comparisons}
\label{app:additional_comparison}

\begin{figure}[!htbp]
  \centering
  \includegraphics[width=.86\linewidth]{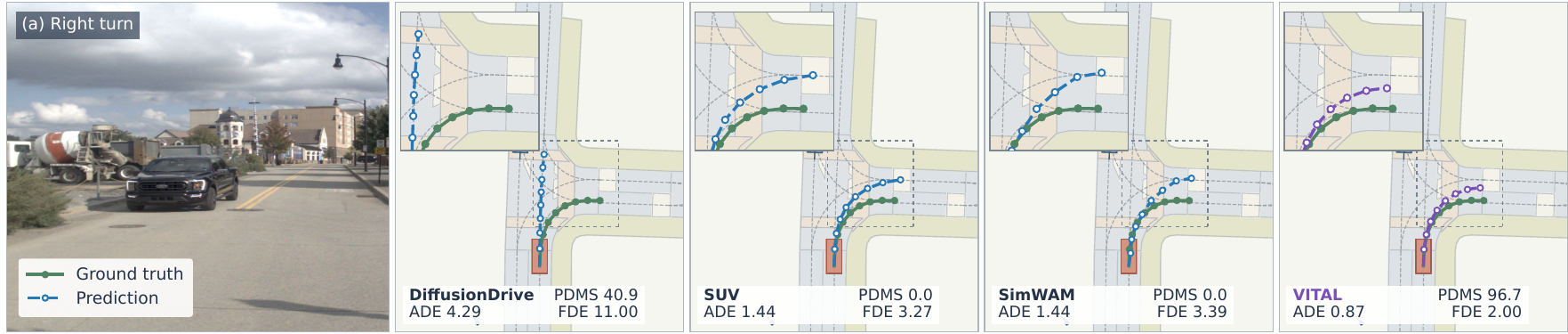}\par\vspace{2pt}
  \includegraphics[width=.86\linewidth]{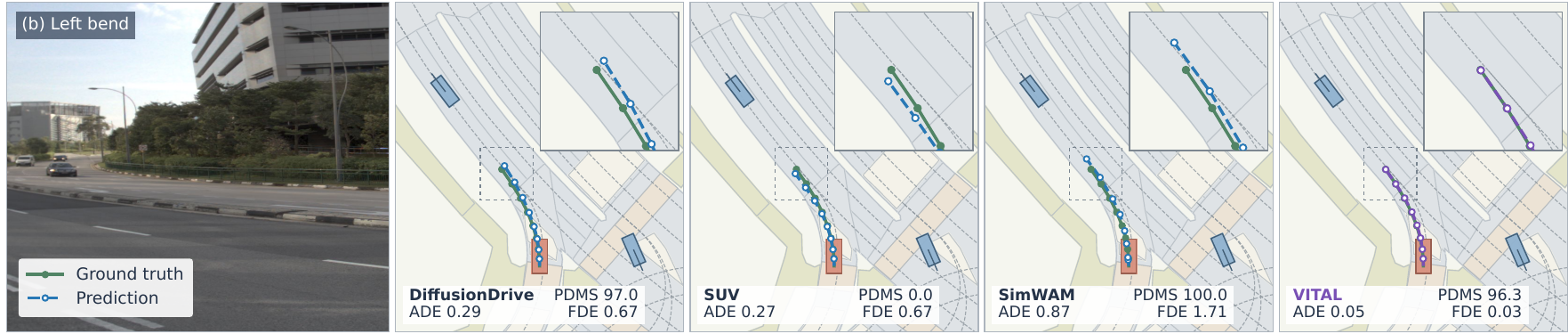}\par\vspace{2pt}
  \includegraphics[width=.86\linewidth]{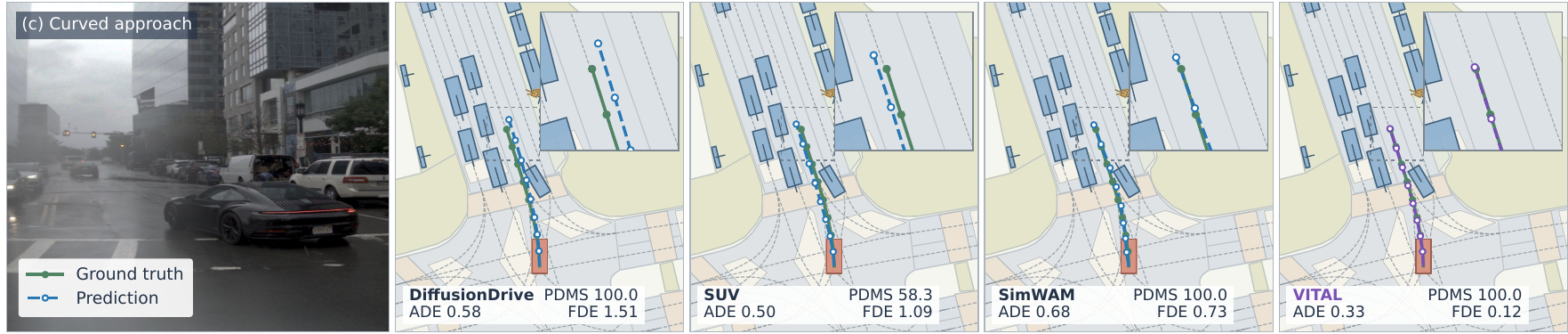}\par\vspace{2pt}
  \includegraphics[width=.86\linewidth]{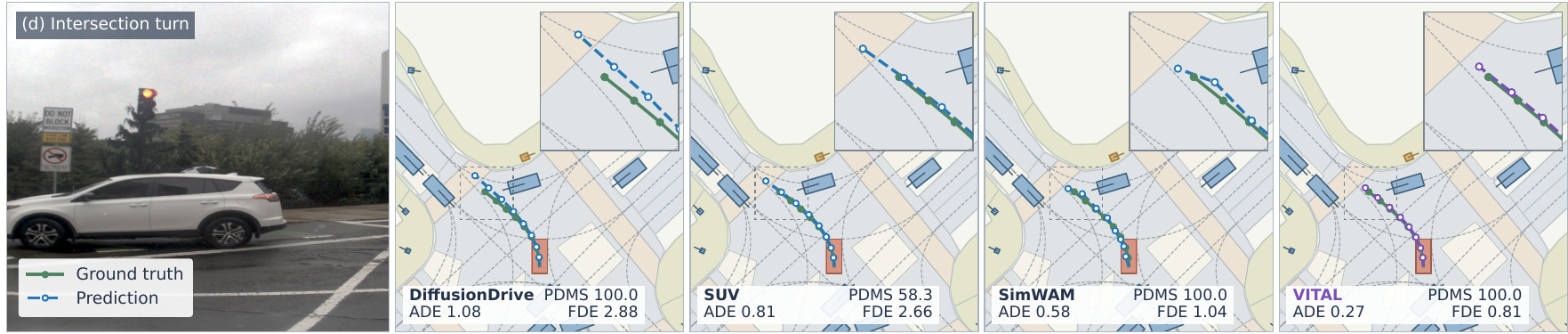}
  \caption{\textbf{Additional trajectory comparisons.} Shared insets show local path differences across four road layouts. ADE and FDE are in meters.}
  \label{fig:vital_additional_comparison}
\end{figure}

Fig.~\ref{fig:vital_additional_comparison} shows smaller endpoint overshoot for \method{} than
SimWAM-RL in the right turn and left bend. In panel b, \method{} ends 0.03\,m from the recorded
endpoint and SimWAM-RL ends 1.71\,m away, while their PDMS values are 96.3 and 100.0.
FDE captures agreement with recorded motion, while PDMS evaluates execution against the PDM reference.

\subsection{Failure Cases and Scope of Residual Refinement}
\label{app:limitations}

\begin{figure}[!htbp]
  \centering
  \includegraphics[width=.50\linewidth]{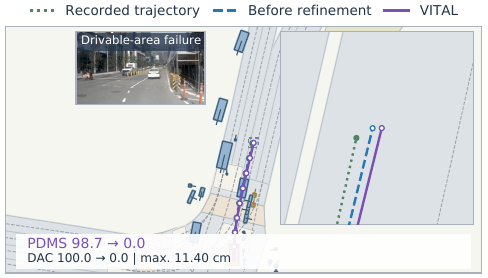}
  \caption{\textbf{Drivable-area boundary crossing.} The inset shows where the refined path leaves the drivable region.}
  \label{fig:vital_failure_cases}
\end{figure}

\independent{\textbf{Boundary sensitivity.}}
In Fig.~\ref{fig:vital_failure_cases}, an 11.40\,cm correction crosses the road boundary.
The DAC gate drops PDMS from 98.7 to 0.0 with no change in NC. Fig.~\ref{fig:vital_cases}
shows the same gate switching in the opposite direction. In the paired one-step evaluation,
64 distinct scenes record a decrease in at least one of four components, NC, DAC, TTC, or DDC.

\independent{\textbf{Local refinement and maneuver choice.}}
The actor adjusts execution along six residual directions with the visual scaffold fixed. The
boundary examples show why precision matters even when the maneuver is unchanged. Reactive
continuations in Appendix~\ref{app:v2_transfer} can call for a different maneuver rather than a local correction to the current scaffold.

\par

\FloatBarrier
\section{Potential Societal Impact \& Limitations}
\label{app:societal_impact}

\subsection{Potential Societal Impact}
\label{app:societal_benefits}
Tab.~\ref{tab:vital_sampling_values} shows that one action evaluation reduces planning latency
in the measured setup. A driving system could use part of that time for independent trajectory
checks. The boundary cases in Appendix~\ref{app:limitations} show why such checks matter even
when the visual scaffold remains fixed and residual corrections move it by only a few centimeters.

\subsection{Known Limitations}
\label{app:known_limitations}
The \method{} action tokenizer has been trained only on autonomous driving data and tasks.
The approach could extend to other vision-driven tasks such as robotic manipulation and embodied navigation.
We plan to pretrain on large collections of paired visual observations and actions, then adapt the tokenizer to the action space of each task.

\section{Public Resources Used}
\label{app:public_resources}
The resources below support training, evaluation, and visualization. The scenes in
Fig.~\ref{fig:vital_teaser} come from public driving footage\footnote{\url{https://www.youtube.com/watch?v=sUcWuiUpvF4&t=9083s}} and illustrate where local precision matters in trajectory planning.

\subsection{Public Datasets Used}
\label{app:public_datasets}
\begin{itemize}
    \item nuPlan~\citep{karnchanachari2024nuplan}\footnote{\url{https://www.nuscenes.org/nuplan}} \dotfill nuPlan Dataset Terms
    \item nuScenes\footnote{\url{https://www.nuscenes.org/nuscenes}} \dotfill CC BY-NC-SA 4.0
    \item OpenScene\footnote{\url{https://github.com/OpenDriveLab/OpenScene}} \dotfill CC BY-NC-SA 4.0
\end{itemize}

\subsection{Public Implementations Used}
\label{app:public_models}
\begin{itemize}
    \item NAVSIM v1/v2\footnote{\url{https://github.com/autonomousvision/navsim}} \dotfill Apache License 2.0
    \item nuPlan-devkit\footnote{\url{https://github.com/motional/nuplan-devkit}} \dotfill Apache License 2.0
    \item VGGT-$\Omega$\footnote{\url{https://github.com/facebookresearch/vggt-omega}} \dotfill FAIR Noncommercial Research License
    \item Depth Anything 3\footnote{\url{https://github.com/ByteDance-Seed/Depth-Anything-3}. Model terms vary.} \dotfill Apache License 2.0 for code
    \item LingBot-Map\footnote{\url{https://github.com/Robbyant/lingbot-map}} \dotfill Apache License 2.0
    \item LingBot-VA / VA 2.0\footnote{\url{https://github.com/Robbyant/lingbot-va}. Includes the VA 2.0 technical report.} \dotfill Apache License 2.0 for code
    \item StarVLA\footnote{\url{https://github.com/starVLA/starVLA}} \dotfill MIT with additional attribution terms
    \item SimWAM\footnote{\url{https://github.com/H-EmbodVis/SimWAM}} \dotfill MIT License
    \item Wan2.2\footnote{\url{https://github.com/Wan-Video/Wan2.2}} \dotfill Apache License 2.0
    \item UMT5-XXL\footnote{\url{https://huggingface.co/google/umt5-xxl}} \dotfill Apache License 2.0
    \item SUV\footnote{\url{https://github.com/ASH-2046/SUV}} \dotfill MIT License
    \item DiffusionDrive\footnote{\url{https://github.com/hustvl/DiffusionDrive}} \dotfill MIT License
    \item DiffSynth-Studio\footnote{\url{https://github.com/modelscope/DiffSynth-Studio}} \dotfill Apache License 2.0
    \item PyTorch\footnote{\url{https://github.com/pytorch/pytorch}} \dotfill BSD 3-Clause License
    \item Transformers\footnote{\url{https://github.com/huggingface/transformers}} \dotfill Apache License 2.0
    \item ROS 2\footnote{\url{https://github.com/ros2/ros2}. Licenses are specified per package.} \dotfill Apache 2.0 / BSD 3-Clause
    \item NumPy\footnote{\url{https://github.com/numpy/numpy}} \dotfill BSD 3-Clause License
\end{itemize}

\clearpage
\bibliographystyle{vitalplainnat}
\bibliography{references}

\end{document}